\pdfoutput=1
\documentclass[11pt]{article}
\usepackage[margin=1in]{geometry}
\usepackage[T1]{fontenc}
\usepackage{lmodern}
\usepackage{microtype}
\usepackage{amsmath,amssymb}
\usepackage{graphicx}
\usepackage{booktabs}
\usepackage{array}
\usepackage{multirow}
\usepackage{makecell}
\usepackage{adjustbox}
\usepackage{enumitem}
\usepackage{xcolor}
\usepackage{tikz}
\usetikzlibrary{positioning,arrows.meta,calc}
\usepackage[numbers,sort&compress]{natbib}
\usepackage[hidelinks]{hyperref}
\usepackage{caption}
\usepackage{placeins}
\usepackage{titlesec}
\titleformat{\paragraph}[runin]{\normalfont\normalsize\bfseries}{}{0pt}{}[.]
\title{Choosing Models for Network-Wide Freeway Speed Nowcasting and Forecasting from Sparse Fixed Sensors: An Evaluation by Sensor Coverage and Network Characteristics}
\author{
  Ran Tu\thanks{Corresponding author: \texttt{rtu@nctcog.org}} \qquad Zihao Li \qquad Arash Mirzaei \\[4pt]
  \normalsize North Central Texas Council of Governments, Arlington, TX 76011, USA \\
  \normalsize\texttt{\{rtu, zli, amirzaei\}@nctcog.org}
}

\date{}

\begin{document}
\maketitle
\begin{abstract}
Most freeway agencies have fixed detectors on only part of their network but need speeds for all of it. This paper
compares methods for estimating current and next-hour speeds on every link from such sparse detectors, using
historical probe data for training. The test network is the Dallas--Fort Worth freeway system: 4{,}127 directional
links, of which 643 (15.6\%) carry a side-fire detector. Eighteen methods from five families (non-learned baselines,
statistical estimators, tabular learners, road-graph neural networks and models that learn spatial relations) were
trained under one protocol with three random seeds. Each neural
architecture was also retrained with a parameter-free, direction-aware kernel interpolation of the sensor readings as
additional input. Graph neural networks that pass information along the road graph, including a plain graph
convolutional recurrent network, reduced the current-hour error on unsensored links by about 17\% relative to the
historical average. Models with adaptive graphs, attention or identity embeddings stayed close to the historical
average and did not fit the training data well, indicating that they could not learn which detectors matter for
which links. With the interpolation features, all ten architectures came within about 0.1~mph of each other. The
remaining error depended mainly on coverage: almost all of the improvement was found within about 3~mi of a detector,
and very little on tollways without detectors or on links with no road-graph connection to one. The results are
summarised as guidance on model choice and detector placement.
\end{abstract}

\section{Introduction}\label{sec:intro}

Regional traffic management relies on knowing speeds across the whole network. Traveller information, dynamic
message signs, incident detection, service patrol dispatch, performance reporting and the calibration of demand and
emission models all use link speeds, including on roads without detectors. Fixed detectors are expensive to install
and maintain, so agencies place them on the busiest urban corridors first. In the Dallas--Fort Worth (DFW) region,
side-fire radar detectors cover 15.6\% of the 4{,}127 directional freeway links. About three quarters of the remaining
links are more than a mile from the nearest detector, one in seven has no road-graph connection to any detector, and
the tollway system has a single detector. Commercial probe data do cover every link, but many agencies buy them in
historical batches for planning and reporting rather than as a real-time feed.

For an agency in this situation, two questions follow. The first is which method to use to estimate the current and
next-hour speed of every link from its own detectors, given that historical probe data can be used to train the
method. The second is where the existing detectors stop being informative, which tells the agency where additional
detectors would help most. If a simple model recovers most of what can be recovered, network-wide speed maps and
congestion alerts can be produced on ordinary hardware without a real-time probe feed. If no model recovers much in
some part of the network, more detectors are needed there.

The literature offers many candidate methods but little guidance for this choice. Graph convolutional recurrent
networks (GCRNs) \citep{seo2018gcrn,li2018dcrnn} and later models with learned graphs and attention
\citep{bai2020agcrn,liu2023staeformer} are widely used for traffic forecasting, but they were developed for detector
networks in which every node has its own time series. Spatio-temporal kriging methods and models for forecasting at
unobserved locations \citep{wu2021ignnk,zheng2023increase,xu2025kits,su2024stsm,zhou2025mogernn} do handle nodes
without input. They are usually tested on dense loop-detector datasets with a few hundred sensors, where the
``unobserved'' nodes are held-out detectors next to observed ones. Statistical estimators and simple interpolation are
seldom part of these comparisons, and the compared models often receive different spatial information, which makes
it hard to tell whether a difference comes from the architecture or from the inputs. We are not aware of a comparison
of these methods on a large, sparsely and unevenly instrumented network, or of an analysis of how their relative
performance depends on local coverage and road characteristics.

This paper provides such a comparison for the DFW freeway network. We evaluate five families of methods: non-learned
baselines, statistical estimators fitted on the probe history, tabular learners, graph neural networks on the road
graph, and models that learn spatial relations from data. The task is to estimate the speed of every link in the
current hour and to forecast it for the next hour. Every neural architecture is trained twice, once in its original
form and once with a parameter-free interpolation of the detector readings along the road network as additional
input. This second version gives all models the same spatial information, so that the effect of the architecture
can be separated from that of the inputs. The contributions are as follows; each corresponds to one of the gaps
discussed in Section~\ref{sec:gaps}.
\begin{enumerate}[nosep,leftmargin=*]
  \item An evaluation on a realistic sparse network (gap G1). The setting, which we call probe-supervised sparse
  sensing, covers all 4{,}127 links of a regional freeway network with 643 instrumented links, uses historical
  probe speeds as training labels, and reports errors by distance to the nearest detector, road type, traffic
  volume, time of day and congestion.
  \item A comparison of 18 methods under one protocol and with matched inputs (gap G2), including statistical
  estimators fitted on the probe history and four models published between 2022 and 2025.
  \item A test of why some graph models fail under sparse sensing (gap G3). We examine three possible causes
  (limited receptive field, sparse inputs, and the difficulty of learning spatial relations from sparse data) in an
  experiment that changes only the spatial operator of one recurrent network. Only the third is supported by the
  results, and the interpolation input removes it for every architecture tested.
  \item Guidance by coverage and network characteristics (gap G4). We relate accuracy to the number of detectors,
  the distance to the nearest detector, road type and traffic volume, report its effect on congestion detection and
  a network travel time index, and summarise the results as recommendations for model choice and detector
  placement.
\end{enumerate}

The detectors are used to define where real-time information is available, and the readings at these locations are
taken from the probe data. The experiments therefore measure how much of the network can be recovered from the
information available at the detector locations, with inputs and targets on the same speed definition.

Section~\ref{sec:related} reviews related work and identifies the research gaps. Section~\ref{sec:method} describes
the data, the methods and the experimental design. Section~\ref{sec:results} presents and discusses the results, and
Section~\ref{sec:conclusion} concludes.

\section{Literature review}\label{sec:related}

\subsection{Network-wide estimation and prediction from sparse sensors}\label{sec:lit-state}
Estimating traffic conditions between and beyond detectors has a long history. Model-based traffic state
estimation reconstructs speeds along a corridor from traffic-flow models and filtering, for example the adaptive
smoothing method of \citet{treiber2002asm} or extended Kalman filters on macroscopic models
\citep{wang2005ekf}, and probe-vehicle measurements can be assimilated in the same framework
\citep{herrera2010lagrangian}. \citet{seo2017tse} review this work. These methods perform well on instrumented
corridors with regularly spaced detectors, but each corridor needs its own calibration, and extending them to a
regional network with long gaps between detectors is difficult. A related planning problem, the sensor location
problem, asks where a limited number of detectors should be placed so that the network state can be inferred
\citep{gentili2012sensor}.

Data-driven methods treat network speeds as a link-by-time matrix or tensor and fill in the missing entries. Examples
are low-rank matrix and tensor completion \citep{tan2013tensor,asif2016matrix}, spatio-temporal cokriging
\citep{bae2018cokriging}, and the Laplacian-regularised low-rank tensor completion of \citet{nie2023letc} for large
networks. If a history is available for every link, as in our case, such estimators can learn the relation between
each link and the detectors directly from that history.

Deep spatio-temporal kriging uses graph neural networks to estimate values where there is no sensor.
IGNNK \citep{wu2021ignnk} trains a diffusion graph network to reconstruct randomly masked sensors. SATCN
\citep{wu2021satcn}, INCREASE \citep{zheng2023increase} and STGNP \citep{hu2023stgnp} extend this with flexible
aggregators, heterogeneous spatial relations and uncertainty estimates. More recent inductive kriging studies deal
with the mismatch between sparse training graphs and dense inference graphs \citep[KITS;][]{xu2025kits},
generalisation to new sensors and to temporal shifts \citep[STA-GANN;][]{li2025stagann}, sparse and noisy graphs
\citep[DarkFarseer;][]{liang2026darkfarseer} and incomplete observations at the observed sensors
\citep[UniSTOK;][]{xie2026unistok}. \citet{lei2024cdste} obtain probabilistic estimates at sensor-free locations with a
conditional diffusion model. DGAE \citep{zhou2025dgae} is the study closest to ours: it estimates network-wide
freeway states from sparse sensors with a graph auto-encoder and physics-guided propagation. Fewer studies forecast
future conditions where no sensor exists. STSM \citep{su2024stsm} and its successor GenCast \citep{su2026gencast} do
so for regions without observations, and MoGERNN \citep{zhou2025mogernn} fills unobserved nodes with a mixture of
graph experts before forecasting with a graph GRU.

Most of these models are tested on METR-LA (207 loop detectors) and PEMS-BAY (325 detectors). In these datasets the
``unobserved'' nodes are 20--50\% of a dense detector network, are close to observed detectors, and have no labels at
any time. Regional networks are different in two ways: coverage is much sparser and more clustered, and historical
probe data provide labels for every link during training.

\subsection{Comparison of method families}\label{sec:lit-compare}
Table~\ref{tab:lit} groups the methods by how they move information from detectors to other links and by what they
assume about the data.

GCRN \citep{seo2018gcrn} combines spectral graph convolution \citep{defferrard2016chebnet} with an LSTM. DCRNN
\citep{li2018dcrnn} uses bidirectional diffusion convolution on a road-distance graph inside a GRU, which allows
upstream and downstream links to be weighted differently. STGCN \citep{yu2018stgcn} uses temporal convolution instead
of recurrence. In all three, information moves a fixed number of hops per layer, and every node is assumed to have an
input.

Graph WaveNet \citep{wu2019gwnet}, MTGNN \citep{wu2020mtgnn} and AGCRN \citep{bai2020agcrn} learn the adjacency
matrix from node embeddings, and AGCRN also learns node-specific parameters. GMAN \citep{zheng2020gman} and
STAEformer \citep{liu2023staeformer} use spatial attention, ImputeFormer \citep{nie2024imputeformer} uses low-rank
attention for imputation, and STID \citep{shao2022stid} showed that identity embeddings combined with an MLP are a
strong baseline. These models can relate any pair of nodes and perform very well on benchmarks where every node is
observed. Their cost on large networks led to linear-complexity designs such as BigST \citep{han2024bigst}, evaluated
on the LargeST benchmark \citep{liu2023largest}.

Kriging and unobserved-location models (IGNNK, INCREASE, KITS, MoGERNN, DGAE) are designed for nodes without input.
They are usually trained with random masking and aggregate only over observed neighbours. Imputation models such as
GRIN \citep{cini2022grin} and SPIN \citep{marisca2022spin} address gaps in otherwise observed series.

Explicit spatial priors are another option. In graph learning with missing node features, \citet{rossi2022featprop}
found that a parameter-free feature propagation step before a graph neural network works well. Kernel
(Nadaraya--Watson) interpolation \citep{nadaraya1964,watson1964} and kriging are the corresponding classical methods
in spatial statistics. In traffic estimation, such priors usually appear as baselines rather than as inputs to a
learned model.

Graph neural networks also have known limitations when information has to travel over many hops. Repeated averaging
makes node states similar \citep{li2018deeper}, and information from distant nodes is compressed when it passes
through a few intermediate nodes \citep{alon2021bottleneck}. Under sparse sensing, much of the information an
unobserved link needs lies several hops away, so these effects could matter. To our knowledge this has not been
tested.

\begin{table}[t]
\centering
\caption{Families of methods for network-wide speed estimation and prediction. ``Unobserved links'': whether the
method is designed to produce estimates for links without input. ``Labels at targets'': whether it requires or uses
historical ground truth at the target links.}
\label{tab:lit}
\small
\setlength{\tabcolsep}{3pt}
\begin{adjustbox}{max width=\linewidth}
\begin{tabular}{@{}p{0.17\linewidth}p{0.2\linewidth}p{0.2\linewidth}cc p{0.27\linewidth}@{}}
\toprule
Family & Examples & Spatial mechanism & \makecell{Unobserved\\links} & \makecell{Labels at\\targets} & Possible limitation under sparse, clustered coverage \\
\midrule
Flow models, filtering & ASM, EKF \citep{treiber2002asm,wang2005ekf} & traffic-flow dynamics & yes & no & per-corridor calibration; difficult to scale to a region \\
Low-rank, kriging & \citep{tan2013tensor,bae2018cokriging,nie2023letc} & correlations in the history & yes & yes & linear; little forecasting ability \\
GCRN family & GCRN, DCRNN, STGCN & $K$-hop graph convolution & no & no & assumes an input at every node; limited receptive field \\
Learned graph, attention & AGCRN, GWNet, STAEformer, STID & learned relations between all nodes & no & no & relations must be learned from sparse inputs \\
Deep kriging & IGNNK, INCREASE, KITS, DGAE & masked graph aggregation & yes & no & tested on dense networks; mostly current state only \\
Unobserved-location forecasting & STSM, MoGERNN & graph experts, contrastive masking & yes & no & aggregation radius; tested on dense networks \\
Explicit propagation & feature propagation, kernel interpolation & fixed diffusion or kernel & yes & no & no learned non-linear corrections \\
\bottomrule
\end{tabular}
\end{adjustbox}
\end{table}

\subsection{Summary and research gaps}\label{sec:gaps}
Existing work covers kriging of the current state at unobserved locations, forecasting at observed locations and,
more recently, forecasting at unobserved locations. It offers scalable graph and attention architectures,
physically based propagation, and statistical estimators that make use of a labelled history. We see four gaps,
which correspond to the contributions listed in Section~\ref{sec:intro}.
\begin{enumerate}[label=\textbf{G\arabic*.},nosep,leftmargin=*]
  \item Realistic coverage. Deep models have been tested on dense detector networks in which unobserved nodes are a
  minority and lie next to observed ones. The situation of many agencies, with 10--20\% coverage, long gaps,
  uninstrumented subnetworks and historical probe labels for every link, has not been studied on a full regional
  network for both current and next-hour speeds.
  \item Comparisons across families with matched inputs. Studies typically compare a new model with a selection of
  others on their own benchmark. Physical, statistical, tabular, road-graph and learned-structure methods have not
  been compared on the same sparsely instrumented network, and published comparisons do not separate the effect of
  the spatial information a model receives from the effect of its architecture.
  \item Which spatial operators work under sparse sensing. The limitations of message passing are known in general,
  but it is not known whether a limited receptive field, missing inputs or the need to learn spatial relations is
  what limits graph neural networks when most nodes have no input, or whether an explicit distance prior along the
  road network removes the limitation.
  \item Guidance by coverage and network characteristics. Results are usually reported as network-wide averages.
  How the choice of model depends on the number of detectors, the distance to the nearest detector, road type and
  traffic volume, and whether accuracy differences matter for products such as congestion detection and travel time
  indices, is rarely examined. An agency needs this information to choose a method and to decide where to add
  detectors.
\end{enumerate}

\section{Data and method}\label{sec:method}

\subsection{Study network and data}\label{sec:data}
The study area is the freeway and tollway system of the Dallas--Fort Worth (DFW) region, Texas. The network has
$N=4{,}127$ directional Traffic Message Channel (TMC) links with a total length of 2{,}222 directional miles (median
link length 0.40~mi), taken from the 2024 TMC network shapefile. Each link has attributes for length, number of
through lanes, directional annual average daily traffic (AADT), functional class and toll status. Of the links,
2{,}146 are on interstates, 561 on tollways and the rest on other freeways and expressways.

\paragraph{Ground truth} Probe-vehicle speeds for the 23 weekdays of October 2024 were exported day by day from the
regional archive of the National Performance Management Research Data Set (NPMRDS) at 15-minute resolution and
averaged to hourly values; 99.6\% of the link--hour cells have data. We use hourly speeds because the operational
question concerns the current and the next hour, and because hourly averages are less affected by the sampling
noise of 15-minute probe speeds on low-volume links.

\paragraph{Fixed detectors} The region's side-fire radar detectors define the sensor set. Of the 929 detectors that
reported data in September 2022, 785 lie on 643 TMC links of the network (15.6\% of the links and 19.2\% of the
mileage); 534 of these links have one detector and 109 have two to four. The other 3{,}484 links (84.4\%) have no
detector. In this study the detectors determine where real-time information is available, and the reading of a
sensor link is taken from the probe speed of that link. Inputs and targets therefore use the same speed definition,
and the experiments measure how much of the network can be recovered from the information available at the
detector locations. Since only the detector locations are used, the fact that the detector and probe archives come
from different years does not affect the experiments.

\paragraph{Detector availability} In practice, detector data are also intermittent. In the side-fire archive for
September 2022 (13 weekdays), the median sensor link reported data in 85\% of the hours; 209 of the 643 links
reported in fewer than 80\% of the hours and 34 in fewer than half. In a typical hour 89\% of the sensor links
reported, in the worst tenth of the hours 71\% or fewer, and on one day a system-wide outage left only 5\% reporting.
The coverage available to a real-time estimator is thus often lower than the nominal 643 links. This is one reason
for the reduced-coverage experiments in Section~\ref{sec:setup}, which show how the methods behave when part of the
detectors is missing, whether through outages or because fewer detectors are installed.

\begin{figure}[t]
  \centering
  \includegraphics[width=0.78\linewidth]{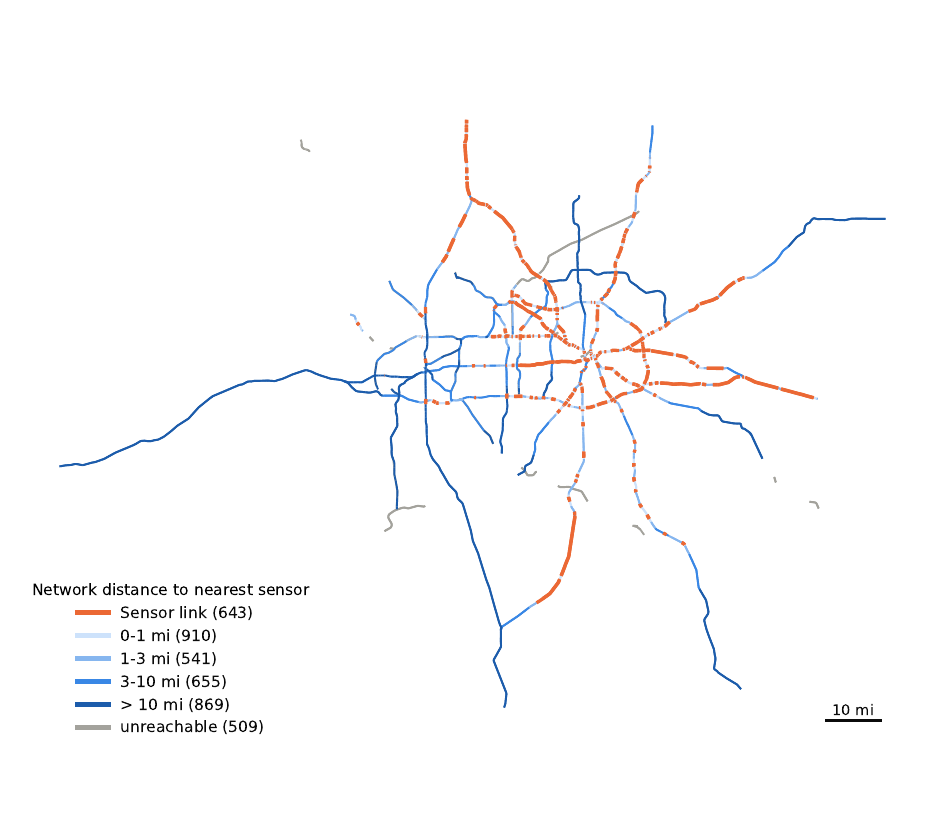}
  \caption{DFW freeway network (4{,}127 directional TMC links). Orange: links with a fixed detector. Blue shades:
  network distance from an unsensored link to the nearest detector (upstream or downstream). Grey: links not
  connected to any detector through the directed road graph. Counts are numbers of links.}
  \label{fig:network}
\end{figure}

\paragraph{Spatial coverage} Figure~\ref{fig:network} shows the network and the network distance from each link to
its nearest detector. Detectors are concentrated on the radial interstates in the urban core, where 20\% of the
interstate links have one; the tollways have a single detector on 561 links. Only 910 unsensored links (26\%) are
within 1~mi of a detector, 869 are more than 10~mi away, and 509 (14.6\%) have no connection to any detector through
the directed road graph described in Section~\ref{sec:graph}. These are mostly tollway and frontage segments, and
some of the missing connections may be due to interchange links that are absent from the TMC geometry. We keep the
graph as built and report these links as a separate group. Because coverage varies so much across the network, the
errors can be related to local coverage.

\subsection{Problem statement}\label{sec:task}
Let $v_{i,t}$ be the speed on link $i$ in hour $t$, $\mathcal S$ the set of sensor links and $m_{i,t}\in\{0,1\}$
an indicator of whether link $i$ has a reading in hour $t$ ($m_{i,t}=0$ for $i\notin\mathcal S$). Given the readings
of the last $T=3$ hours, $\{v_{j,t'} : m_{j,t'}=1,\; t-T<t'\le t\}$, a model returns for every link the current speed
$\hat v_{i,t}$ and the next-hour speed $\hat v_{i,t+1}$. We refer to the first as nowcasting (for $i\notin\mathcal S$
it is a spatio-temporal kriging problem) and to the second as forecasting. Probe speeds on all links are used as
labels during training, while at inference only the detector readings are available. We call this setting
probe-supervised sparse sensing. It corresponds to an agency that licenses historical probe data for planning and
operates in real time with its own detectors. It differs from the inductive kriging benchmarks used in recent studies
\citep{wu2021ignnk,zheng2023increase,xu2025kits,zhou2025mogernn}, in which the unobserved locations never have labels.

The days are split in time order into 15 training days (1--21 October), 4 validation days (22--25 October) and 4
test days (28--31 October). The test days are not used in training and cover all 24 hours. Normalisation statistics
and the historical profile are computed from the training days only.

\paragraph{Anomaly decomposition} Speeds are split into a historical profile and a standardised anomaly,
\begin{equation}\label{eq:anomaly}
  v_{i,t} = \mathrm{HA}_i(\tau_t) + \sigma_r\, r_{i,t},
\end{equation}
where $\tau_t$ is the hour of day of hour $t$, $\mathrm{HA}_i(\tau)$ is the mean speed of link $i$ at hour of day
$\tau$ on the training days, and $\sigma_r=4.75$~mph is the standard deviation of the training anomalies
$v-\mathrm{HA}$. The profile already contains recurrent congestion. What remains to be estimated is the deviation $r$
of the current day, which is observed only at the detectors. All models, including the baselines, predict $r$; errors
are reported in mph.

\subsection{Directed road graph}\label{sec:graph}
The nodes of the graph are directional TMC links. A directed edge $i\to j$ is added when the downstream end of $i$ is
within 100~m of the upstream end of $j$, $j$ is not the opposite carriageway of $i$, and the turning angle is less than
$120^\circ$. This connects consecutive links of the same road and the links of crossing freeways at interchanges
(4{,}399 edges). The row-normalised adjacency matrices $\tilde A_{\downarrow}$ and $\tilde A_{\uparrow}$ average over
the direct successors and the direct predecessors of a link. Let $d_{ij}$ be the shortest directed network distance
from the midpoint of $i$ to the midpoint of $j$, computed up to 15~mi; it is finite only if $j$ is downstream of $i$.
With the kernel $\kappa_h(d)=\exp\bigl(-(d/h)^2\bigr)$ we define two row-normalised distance-kernel transition
matrices with $h=1.5$~mi, truncated at 3~mi:
\begin{equation}\label{eq:transition}
  [P_{\downarrow}]_{ij} \propto \kappa_{1.5}(d_{ij}), \qquad [P_{\uparrow}]_{ij} \propto \kappa_{1.5}(d_{ji}).
\end{equation}
$(P_{\downarrow}X)_i$ is a weighted average of $X$ over the links up to 3~mi downstream of $i$, and $(P_{\uparrow}X)_i$
the corresponding average over the links up to 3~mi upstream.

\subsection{Methods compared}\label{sec:families}
The methods fall into five families, which differ mainly in how information is passed from the detectors to other
links (Table~\ref{tab:lit}). All learned models receive the same inputs for each input hour: the masked anomaly
$m_{i,t'}r_{i,t'}$, the mask $m_{i,t'}$, the mean anomaly $\bar r_{t'}$ over all reporting detectors (which also
reaches links without a road-graph connection to a detector), $\mathrm{HA}_i(\tau_{t'})$, the hour of day as sine and
cosine, and seven static attributes (lanes, log AADT, length, interstate and toll flags, and the daily mean and
minimum of $\mathrm{HA}_i$). All of them predict the anomaly of the current and of the next hour.
\begin{enumerate}[label=(\alph*),nosep,leftmargin=*]
  \item Non-learned baselines. HA is the historical average. HA + interpolation adds the interpolation baseline $b$
  of Eq.~\eqref{eq:baseline} to HA and keeps the current anomaly for $t{+}1$. Persistence of the last reading
  ($\hat v_{t+1}=v_t$) is reported for sensor links only.
  \item Statistical estimators fitted on the probe history. Covariance kriging estimates the covariance of the
  anomalies of all links from the training days, shrinks it towards its diagonal \citep{ledoit2004} and computes
  the Gaussian conditional expectation given the detector readings. Low-rank (PPCA) does the same with a
  probabilistic principal component model \citep{tipping1999ppca}. Rank, shrinkage and noise level are selected on
  the validation days.
  \item Tabular learners. Ridge regression and histogram gradient boosting (GBDT) in the style of LightGBM
  \citep{ke2017lightgbm} are fitted on a per-link feature table with the common inputs and the interpolation features
  of Section~\ref{sec:interp}.
  \item Graph neural networks on the road graph. The graph convolutional recurrent network (GCRN) of
  \citet{seo2018gcrn} replaces the matrix products in a recurrent cell by graph convolutions. We use the diffusion
  convolution of DCRNN \citep{li2018dcrnn} inside a GRU \citep{cho2014gru},
  \begin{equation}\label{eq:diffconv}
    \Theta\star_{\mathcal G} X = X\Theta_{0} + \sum_{k=1}^{K}\bigl(S_{\downarrow}^{k}X\,\Theta_{k}^{\downarrow}
    +S_{\uparrow}^{k}X\,\Theta_{k}^{\uparrow}\bigr),
  \end{equation}
  where $X\in\mathbb R^{N\times C}$ contains the input and hidden state of all links and each $\Theta$ is a weight
  matrix; the network has two layers, 32 hidden units and $K=2$. The model labelled GCRN uses the 1-hop adjacency
  ($S=\tilde A$), the usual graph for this type of model, and the model labelled DCRNN uses the distance-kernel
  graph ($S=P$). IGNNK \citep{wu2021ignnk} (three diffusion graph convolution layers over the input window), MoGERNN
  \citep{zhou2025mogernn} (a mixture of distance-weighted graph experts that fills unobserved links, followed by a
  diffusion-convolutional GRU) and DGAE \citep{zhou2025dgae} (a graph auto-encoder for network-wide freeway
  estimation) also aggregate over road-graph neighbours with distance weights.
  \item Models with learned spatial structure. These models learn which links are related instead of taking this
  from the road graph. AGCRN \citep{bai2020agcrn} is a GCRN with a learned adaptive graph and node-specific
  parameters. STAEformer \citep{liu2023staeformer} combines adaptive embeddings with spatial attention, ImputeFormer
  \citep{nie2024imputeformer} uses low-rank spatial attention, and STID \citep{shao2022stid} uses identity embeddings
  and an MLP without message passing. A per-link GRU with no spatial operator, which knows about other links only
  through $\bar r$, serves as a lower reference.
\end{enumerate}

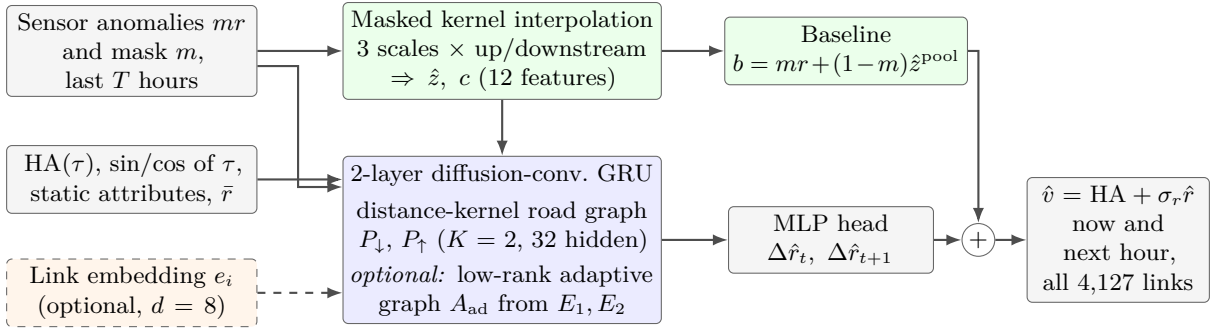
\begin{figure}[t]
  \centering
  \begin{adjustbox}{max width=\linewidth}\begin{tikzpicture}[
    font=\footnotesize,
    box/.style={draw=black!60, rounded corners=2pt, align=center, inner sep=3pt},
    inp/.style={box, fill=black!4, text width=31mm},
    arr/.style={-{Latex[length=2mm]}, black!70, thick}]
  \node[inp] (sens) at (0, 0.2) {Sensor anomalies $m r$\\ and mask $m$, last $T$ hours};
  \node[inp] (ctx)  at (0,-1.5) {HA$(\tau)$, $\sin$/$\cos$ of $\tau$,\\ static attributes, $\bar r$};
  \node[box, dashed, fill=orange!8, text width=31mm] (emb) at (0,-3.0) {Link embedding $e_i$\\ (optional, $d=8$)};
  \node[box, fill=green!8, text width=40mm] (itp) at (4.9, 0.2)
     {Masked kernel interpolation\\ 3 scales $\times$ up/downstream\\ $\Rightarrow \hat z,\ c$ (12 features)};
  \node[box, fill=blue!8, text width=40mm, minimum height=22mm] (gru) at (4.9,-2.3)
     {2-layer diffusion-conv.\ GRU\\[3pt] distance-kernel road graph\\ $P_{\downarrow}$, $P_{\uparrow}$ ($K=2$, 32 hidden)\\[2pt] \textit{optional:} low-rank adaptive\\ graph $A_{\mathrm{ad}}$ from $E_1,E_2$};
  \node[box, fill=green!8, text width=30mm] (base) at (9.45, 0.2) {Baseline\\ $b=m r+(1-m)\hat z^{\mathrm{pool}}$};
  \node[box, fill=black!4, text width=25mm] (head) at (9.25,-2.3) {MLP head\\ $\Delta\hat r_t,\ \Delta\hat r_{t+1}$};
  \node[circle, draw=black!60, inner sep=1pt] (plus) at (11.2,-2.3) {$+$};
  \node[box, fill=black!4, text width=22mm] (out) at (13.05,-2.3) {$\hat v=\mathrm{HA}+\sigma_r\hat r$\\ now and next hour,\\ all 4{,}127 links};
  \draw[arr] (sens) -- (itp);
  \draw[arr] ([yshift=-2mm]sens.east) -- (2.2,0.0) |- ([yshift=7mm]gru.west);
  \draw[arr] (ctx.east) -- (ctx.east -| gru.west);
  \draw[arr, dashed] (emb.east) -- ([yshift=-7mm]gru.west);
  \draw[arr] (itp) -- (gru);
  \draw[arr] (itp) -- (base);
  \draw[arr] (gru) -- (head);
  \draw[arr] (head) -- (plus);
  \draw[arr] (base.east) -| (plus.north);
  \draw[arr] (plus) -- (out);
\end{tikzpicture}\end{adjustbox}
  \caption{Interpolation input combined with a graph recurrent network (GCRN+). The detector anomalies are spread
  over the network by a parameter-free, direction-aware kernel interpolation, which provides input features and an
  output baseline. A two-layer diffusion-convolutional GRU on the road graph learns corrections to the baseline for
  the current and the next hour. Dashed: the link-specific components of GCRN+.}
  \label{fig:arch}
\end{figure}

\paragraph{Link-specific components} Because probe labels exist for every link, parameters specific to each link can
be trained. To see whether they help, we include GCRN+, a DCRNN with the interpolation input and two such components:
a low-rank adaptive graph similar to those of Graph WaveNet and AGCRN \citep{wu2019gwnet,bai2020agcrn},
\begin{equation}\label{eq:adaptive}
  A_{\mathrm{ad}} X = \frac{\mathrm{ReLU}(E_1)\bigl(\mathrm{ReLU}(E_2)^{\top}X\bigr)}{\mathrm{ReLU}(E_1)\bigl(\mathrm{ReLU}(E_2)^{\top}\mathbf 1\bigr)},
  \qquad E_1,E_2\in\mathbb R^{N\times q},\quad q=10,
\end{equation}
used as a third support in Eq.~\eqref{eq:diffconv} (the division is row-wise), and an 8-dimensional learnable
embedding for each link, appended to the input at every step (Figure~\ref{fig:arch}). The low-rank form costs
$O(NqC)$ per application; the dense form $\mathrm{softmax}(\mathrm{ReLU}(E_1E_2^{\top}))$ would have
$1.7\times10^7$ entries. AGCRN uses the same low-rank form in our implementation.

\subsection{Interpolation input}\label{sec:interp}
The families differ in whether they know the road distance between a detector and a link. To separate this
knowledge from the architecture, we use a parameter-free interpolation that can be given to any model. For each
scale $s\in\{1,3,8\}$~mi, a downstream kernel $W^{(s,\downarrow)}_{ij}=\kappa_s(d_{ij})$ and an upstream kernel
$W^{(s,\uparrow)}_{ij}=\kappa_s(d_{ji})$ are defined and truncated at $\min(3s,15)$~mi. For each of the six kernels
$W$ we compute
\begin{equation}\label{eq:interp}
  \hat z_{i,t}=\frac{\sum_j W_{ij}\, m_{j,t}\, r_{j,t}}{\sum_j W_{ij}\, m_{j,t}+\lambda},\qquad
  c_{i,t} = \log\Bigl(1+\sum_j W_{ij}\, m_{j,t}\Bigr),
\end{equation}
a Nadaraya--Watson estimate of the anomaly \citep{nadaraya1964,watson1964} and a measure of how much detector data
it is based on. The constant $\lambda>0$ shrinks the estimate towards zero, i.e.\ towards the historical profile,
where there is little detector data. With $\lambda=10^{-3}$, the three scales and two directions give 12 features
per link and hour. The same estimate with the pooled kernel $W^{(1,\downarrow)}+W^{(1,\uparrow)}$ and $\lambda=0.1$
(both chosen on the validation days), $\hat z^{\mathrm{pool}}$, gives an output baseline
\begin{equation}\label{eq:baseline}
  b_{i,t} = m_{i,t}\, r_{i,t} + (1-m_{i,t})\,\hat z^{\mathrm{pool}}_{i,t},
\end{equation}
so that a model with the interpolation input learns a correction to an interpolated first estimate. The two
directions are kept separate because disturbances travel upstream in congestion and downstream in free flow
\citep{lighthill1955kinematic,richards1956shock}. The interpolation costs two sparse matrix products per kernel.
Every learned architecture except DGAE was trained without and with it (12 features added to its inputs at every
step and $b$ added to its output); with the interpolation input, all models have the same inputs. Ridge and GBDT
always use the features, and DCRNN + interp.\ denotes DCRNN with the interpolation input.

\subsection{Three possible limitations of graph models under sparse sensing}\label{sec:why}
Graph neural networks for traffic were developed for detector networks in which every node has an input. Here 84\%
of the nodes have none, and the spatial operator may limit accuracy in three ways. We state these as hypotheses and
test them in Section~\ref{sec:res-why}.

\begin{figure}[t]
  \centering
  \includegraphics[width=0.5\linewidth]{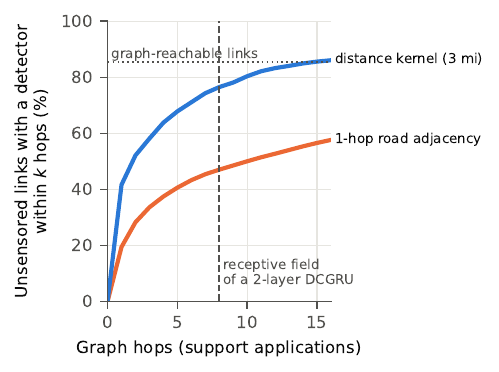}
  \caption{Share of the 3{,}484 unsensored links with a detector within $k$ applications of the graph operator. The
  dashed line marks the receptive field of the two-layer GRU used here over three input hours ($K(T+L-1)=8$). The
  dotted line is the share of links connected to any detector through the directed road graph (85.4\%).}
  \label{fig:reach}
\end{figure}

H1, limited receptive field. Each convolution moves information by at most $K$ hops, so in a GRU with $L$ layers run
over $T$ input steps the top-layer state of a link depends only on inputs within $K(T+L-1)=8$ hops. With the 1-hop
adjacency, 47\% of the unsensored links have a detector within 8 hops; with the distance-kernel graph, where one hop
can cover 3~mi, the share is 77\% (Figure~\ref{fig:reach}). A deeper network would reach further but would also
average out more of the signal \citep{li2018deeper,alon2021bottleneck}.

H2, diluted inputs. Unsensored links enter the convolution with $m r=0$. The aggregated input of a link is then the
weighted mean anomaly of the detectors around it multiplied by their share of the weights, which has a median of
0.34 on the distance-kernel graph. To recover the anomaly, the network has to approximate a division with its gates,
unless it receives a normalised estimate such as Eq.~\eqref{eq:interp}. Feature propagation addresses the same issue
for graphs with missing node features \citep{rossi2022featprop}.

H3, relations that must be learned. Adaptive graphs, attention and identity embeddings have to learn from data which
of the 643 detectors are relevant to each of the 4{,}127 links. Most inputs are empty, the training period has 15
days, and the gradient that shapes the learned relations passes only through the sensor nodes. These models are not
told that nearby links on the same road are related.

\subsection{Experimental design}\label{sec:setup}

\paragraph{Training} All deep models use the same training procedure. Samples are sliding windows within a day (21
per day, 315 training windows, each covering all 4{,}127 links). In each sample a random fraction $U(0,0.4)$ of the
detectors is hidden, i.e.\ set to $m=0$ in all inputs including the interpolation, so that the models learn to
reconstruct sensor links from the remaining detectors. The loss is a Huber loss ($\delta=1$) on standardised
anomalies over all links with ground truth; current-hour cells of observed detectors receive a weight of 0.1 because
their value is an input. We use Adam with a learning rate of $2\times10^{-3}$, halved after 2 epochs without
improvement on the validation set, a batch size of 8, gradient clipping at 5 and at most 40 epochs, with early
stopping after 6 epochs without improvement in the validation RMSE of unsensored links. Hyperparameters are listed
in Appendix~\ref{app:hyper}.

\paragraph{Adaptations} Several models had to be adapted to the task and to CPU training.
(a)~All models output the current and the next-hour anomaly; output heads designed for one horizon (IGNNK) or for
long horizons (STAEformer, MoGERNN) were reduced accordingly.
(b)~Spatial self-attention over 4{,}127 tokens took 42~s per batch for STAEformer on our CPU. We therefore let every
link act as a query and only the 643 sensor links act as keys and values. Only these tokens carry information about
the current day, so STAEformer can still estimate unobserved links in the same way. Attention uses a
memory-efficient exact implementation \citep{dao2022flashattention}.
(c)~ImputeFormer's spatial attention uses the linear form $\mathrm{softmax}(Q)\,(\mathrm{softmax}(K)^{\top}V)$
\citep{shen2021efficientattention}. Its Fourier sparsity regulariser is not meaningful for a 3-step window and was
left out.
(d)~MoGERNN's experts aggregate over the 48 nearest links within 5~mi network distance, as in the original paper, and
only read neighbours that have a reading.
(e)~All models use the common sensor dropout instead of their own masking schemes.
(f)~In its original configuration (post-LayerNorm and Xavier-initialised adaptive embedding
\citep{glorot2010understanding}), STAEformer produced the same prediction for every link and was no better than HA
(Appendix~\ref{app:stae}). We report it with pre-LayerNorm blocks \citep{xiong2020layernorm}, $\mathcal N(0,0.1^2)$
initialisation of the adaptive embedding and its original learning rate of $10^{-3}$.
The results therefore refer to the adapted models on this task and not to the published configurations on their
original tasks.

\paragraph{Coverage scenarios and stratification} We vary coverage in two ways. For global coverage, representative
models of each family were retrained after removing half or three quarters of the sensor links at random (322 and
161 detectors, 7.8\% and 3.9\% of the links); the removed sensor links are added to the evaluation set. Random removal
mimics scattered outages or a thinner deployment, but not correlated outages of whole corridors or of the whole
system (Section~\ref{sec:limitations}). For local coverage and network characteristics, the errors at full coverage
are broken down by network distance to the nearest detector (0--1, 1--3, 3--10 and over 10~mi, and not connected), by
road type (interstate, other freeway, tollway) and by directional AADT tercile. We also report errors by period of
the issue hour $t$ (AM peak 6:00--9:00, PM peak 15:00--19:00, other hours) and for congested cells (true speed below
45~mph in the target hour).

\paragraph{Metrics and repetitions} MAE and RMSE (mph) are computed over the test days for issue hours
$t=2,\dots,22$ and over cells with ground truth; the first two hours of a day have no complete input window, and hour
23 has no next-hour target on the same day. The main target is the set of unsensored links (291{,}694 cells per
horizon). Sensor links are evaluated at $t{+}1$ only, because their current speed is an input. To compare two models
hour by hour, we report the share of the 84 test hours in which each has the lower hourly MSE on unsensored links;
because the hours come from four days and are correlated, these shares are descriptive only. All learned models were
trained with three random seeds (42, 43 and 44), except STAEformer and its variant with the interpolation input
(about 12 hours per run on our CPU), the intermediate ablation variants and the reduced-coverage runs, which use
seed 42. Tables give the mean and, where available, the standard deviation over seeds. The code is written in
PyTorch without graph libraries and runs on a 10-core CPU.

\section{Results and discussion}\label{sec:results}

\begin{table}[t]
\centering
\caption{Test-day accuracy (mph). Unsensored links: 3{,}484 links, 291{,}694 cells per horizon. Sensor links:
next hour only. Mean over random seeds $\pm$ standard deviation (the last column gives the number of seeds; the
statistical estimators are deterministic). Best value per column in bold. Ridge, GBDT, DCRNN + interp.\ and GCRN+
use the interpolation features; the other models are shown in their original form (their variants with
interpolation features are in Table~\ref{tab:featmatch}). The GCRN family differs only in the spatial operator:
none, 1-hop road adjacency, distance-kernel road graph, or a learned graph.}
\label{tab:main}
\small
\setlength{\tabcolsep}{4pt}
\begin{adjustbox}{max width=\linewidth}
\begin{tabular}{lrccccc c}
\toprule
 & & \multicolumn{2}{c}{Unsensored, now ($t$)} & \multicolumn{2}{c}{Unsensored, next hour ($t{+}1$)} & Sensor, $t{+}1$ & \\
\cmidrule(lr){3-4}\cmidrule(lr){5-6}\cmidrule(lr){7-7}
Model & Params & RMSE & MAE & RMSE & MAE & RMSE & Seeds \\
\midrule
HA & -- & 5.24 & 2.77 & 5.24 & 2.75 & 6.08 & -- \\
HA + interpolation & -- & 4.48 & 2.32 & 5.18 & 2.79 & 5.94 & -- \\
\midrule
Covariance kriging \citep{ledoit2004} & -- & 5.03 & 2.70 & 5.22 & 2.82 & 5.91 & -- \\
Low-rank (PPCA) \citep{tipping1999ppca} & -- & 4.95 & 2.63 & 5.12 & 2.71 & 5.77 & -- \\
\midrule
Ridge (interp.\ features) & -- & 4.44{\scriptsize$\pm$0.00} & 2.34{\scriptsize$\pm$0.00} & 4.95{\scriptsize$\pm$0.00} & 2.64{\scriptsize$\pm$0.00} & 5.14{\scriptsize$\pm$0.00} & 3 \\
GBDT (interp.\ features) & -- & 4.36{\scriptsize$\pm$0.00} & 2.24{\scriptsize$\pm$0.00} & 4.96{\scriptsize$\pm$0.00} & 2.62{\scriptsize$\pm$0.00} & 5.22{\scriptsize$\pm$0.01} & 3 \\
\midrule
GRU (no graph) & 14k & 5.04{\scriptsize$\pm$0.01} & 2.55{\scriptsize$\pm$0.00} & 5.11{\scriptsize$\pm$0.01} & 2.63{\scriptsize$\pm$0.00} & 5.13{\scriptsize$\pm$0.02} & 3 \\
GCRN (1-hop graph) \citep{seo2018gcrn} & 56k & 4.33{\scriptsize$\pm$0.03} & 2.15{\scriptsize$\pm$0.01} & 4.94{\scriptsize$\pm$0.02} & 2.55{\scriptsize$\pm$0.01} & 5.09{\scriptsize$\pm$0.02} & 3 \\
DCRNN (kernel graph) \citep{li2018dcrnn} & 56k & 4.32{\scriptsize$\pm$0.02} & 2.16{\scriptsize$\pm$0.02} & 4.94{\scriptsize$\pm$0.02} & 2.56{\scriptsize$\pm$0.02} & 5.10{\scriptsize$\pm$0.03} & 3 \\
AGCRN (learned graph) \citep{bai2020agcrn} & 253k & 5.10{\scriptsize$\pm$0.02} & 2.54{\scriptsize$\pm$0.01} & 5.13{\scriptsize$\pm$0.02} & 2.61{\scriptsize$\pm$0.01} & 5.16{\scriptsize$\pm$0.02} & 3 \\
\midrule
IGNNK \citep{wu2021ignnk} & 49k & 4.31{\scriptsize$\pm$0.03} & 2.16{\scriptsize$\pm$0.03} & 4.92{\scriptsize$\pm$0.00} & 2.55{\scriptsize$\pm$0.01} & 5.13{\scriptsize$\pm$0.02} & 3 \\
\midrule
STID \citep{shao2022stid} & 233k & 5.06{\scriptsize$\pm$0.01} & 2.51{\scriptsize$\pm$0.02} & 5.15{\scriptsize$\pm$0.02} & 2.60{\scriptsize$\pm$0.02} & 5.23{\scriptsize$\pm$0.04} & 3 \\
STAEformer \citep{liu2023staeformer} & 1942k & 5.07 & 2.53 & 5.11 & 2.57 & 5.15 & 1 \\
ImputeFormer \citep{nie2024imputeformer} & 535k & 5.09{\scriptsize$\pm$0.02} & 2.51{\scriptsize$\pm$0.00} & 5.11{\scriptsize$\pm$0.03} & 2.58{\scriptsize$\pm$0.01} & 5.14{\scriptsize$\pm$0.03} & 3 \\
MoGERNN \citep{zhou2025mogernn} & 54k & 4.33{\scriptsize$\pm$0.03} & 2.15{\scriptsize$\pm$0.02} & \textbf{4.92{\scriptsize$\pm$0.01}} & 2.54{\scriptsize$\pm$0.01} & \textbf{5.08{\scriptsize$\pm$0.00}} & 3 \\
DGAE \citep{zhou2025dgae} & 63k & 4.34{\scriptsize$\pm$0.01} & 2.15{\scriptsize$\pm$0.00} & 4.96{\scriptsize$\pm$0.01} & 2.56{\scriptsize$\pm$0.01} & 5.22{\scriptsize$\pm$0.02} & 3 \\
\midrule
DCRNN + interp. & 62k & 4.30{\scriptsize$\pm$0.01} & 2.14{\scriptsize$\pm$0.01} & 4.92{\scriptsize$\pm$0.01} & 2.56{\scriptsize$\pm$0.01} & 5.14{\scriptsize$\pm$0.02} & 3 \\
GCRN+ & 207k & \textbf{4.27{\scriptsize$\pm$0.02}} & \textbf{2.09{\scriptsize$\pm$0.01}} & \textbf{4.92{\scriptsize$\pm$0.00}} & \textbf{2.53{\scriptsize$\pm$0.00}} & 5.13{\scriptsize$\pm$0.02} & 3 \\
\bottomrule
\end{tabular}
\end{adjustbox}
\end{table}

\subsection{Comparison of the method families}\label{sec:res-overall}
Table~\ref{tab:main} gives the test-day accuracy at full coverage. On the 3{,}484 unsensored links, HA has an RMSE of
5.24~mph at both horizons. The interpolation baseline, which involves no training, lowers the current-hour RMSE to
4.48~mph (14\% below HA) but hardly improves the next hour (5.18~mph).

The learned models fall into two groups. The first group consists of models that use road distances between
detectors and links, either through a road-graph operator (GCRN, DCRNN, IGNNK, MoGERNN, DGAE) or through the
interpolation features (Ridge, GBDT, DCRNN + interp., GCRN+). Their current-hour RMSE is 4.27--4.44~mph, 15--19\%
below HA. The second group consists of models whose spatial relations are learned (AGCRN, STID, STAEformer,
ImputeFormer) or absent (GRU). Their current-hour RMSE is 5.04--5.10~mph, which is closer to HA than to the
interpolation baseline. Differences within the first group are small. The road-graph neural networks are within
0.08~mph of each other for the current hour (4.27--4.34~mph) and within 0.05~mph for the next hour (4.92--4.96~mph),
with seed-to-seed standard deviations of 0.01--0.03~mph. The plain GCRN on the 1-hop adjacency is as accurate as DCRNN
and as the recent models designed for unobserved locations. GBDT with the interpolation features trains in about a
minute and is within 0.09~mph of the best model for the current hour and within 0.04~mph for the next hour. The two
statistical estimators fitted on the probe history (covariance kriging, 5.03~mph; low-rank PPCA, 4.95~mph) are better
than HA but worse than the interpolation baseline. With only 15 training days, the covariance of all links has to be
shrunk heavily and does not capture the relation between individual links and the detectors.

Forecasting the next hour gains much less than nowcasting; the best models are 6\% below HA. At sensor links, where
the current speed is known, the next-hour RMSE of the road-graph models is 5.08--5.22~mph, compared with 7.67~mph for
persistence of the last reading and 6.08~mph for HA. Hourly speeds change sharply when congestion sets in or clears,
so neither the last value nor the profile is a good forecast. The detectors mainly tell the models about the current
anomaly, and at the hourly scale this anomaly persists only weakly into the next hour.

\subsection{Which spatial operators work under sparse sensing}\label{sec:res-why}
\begin{table}[t]
\centering
\caption{Effect of the spatial operator under sparse sensing: the same recurrent network (two layers, 32 hidden units,
$K=2$, three input hours) with four spatial operators, trained without and with the interpolation input. Reach: share
of unsensored links that have at least one sensor within the receptive field of the network (8 applications of the
graph operator); the learned graph connects every pair of links. RMSE in mph on unsensored links (mean $\pm$ s.d.\ over
seeds). Train loss: Huber loss on standardised anomalies at the epoch with the best validation RMSE (all links,
with sensor dropout), mean over seeds.}
\label{tab:why}
\small
\setlength{\tabcolsep}{4pt}
\begin{adjustbox}{max width=\linewidth}
\begin{tabular}{l c ccc ccc}
\toprule
 & & \multicolumn{3}{c}{Without interpolation} & \multicolumn{3}{c}{With interpolation} \\
\cmidrule(lr){3-5}\cmidrule(lr){6-8}
Spatial operator & Reach & RMSE $t$ & RMSE $t{+}1$ & Train loss & RMSE $t$ & RMSE $t{+}1$ & Train loss \\
\midrule
None (per-link GRU) & 0\% & 5.04{\scriptsize$\pm$0.01} & 5.11{\scriptsize$\pm$0.01} & 0.244 & 4.29{\scriptsize$\pm$0.01} & 4.95{\scriptsize$\pm$0.01} & 0.218 \\
1-hop road adjacency (GCRN) & 47\% & 4.33{\scriptsize$\pm$0.03} & 4.94{\scriptsize$\pm$0.02} & 0.209 & 4.30{\scriptsize$\pm$0.01} & 4.96{\scriptsize$\pm$0.00} & 0.208 \\
Distance-kernel road graph (DCRNN) & 77\% & 4.32{\scriptsize$\pm$0.02} & 4.94{\scriptsize$\pm$0.02} & 0.209 & 4.30{\scriptsize$\pm$0.01} & 4.92{\scriptsize$\pm$0.01} & 0.206 \\
Learned low-rank graph (AGCRN) & all & 5.10{\scriptsize$\pm$0.02} & 5.13{\scriptsize$\pm$0.02} & 0.225 & 4.29{\scriptsize$\pm$0.01} & 4.94{\scriptsize$\pm$0.02} & 0.198 \\
\bottomrule
\end{tabular}
\end{adjustbox}
\end{table}

Table~\ref{tab:why} tests the three hypotheses of Section~\ref{sec:why}. The same recurrent network is trained with
four spatial operators, each without and with the interpolation input.

The limited receptive field (H1) did not reduce accuracy. Within its receptive field, the 1-hop GCRN can reach a
detector from only 47\% of the unsensored links, compared with 77\% for the distance-kernel DCRNN, yet the two models
have the same error (4.33 and 4.32~mph for the current hour, 4.94~mph for the next hour). The distance breakdown in
Table~\ref{tab:distance} explains why. The 1-hop GCRN is somewhat worse between 1 and 3~mi from a detector (4.96 vs.\
4.83~mph) and slightly better within 1~mi (2.98 vs.\ 3.06~mph), and beyond 3~mi neither model, nor any other, does much
better than HA. The links that the shorter receptive field misses are links about which the detectors say little.

Dilution of the inputs (H2) did not reduce accuracy either. If it did, the normalised interpolation features would
help the road-graph models. They change the current-hour RMSE of the 1-hop GCRN and of DCRNN by 0.03~mph or less, and
the training losses stay the same (0.208 and 0.206, against 0.209 without). Since the mask is one of the inputs, the
network can apparently learn the normalisation itself.

The difficulty of learning spatial relations (H3) is consistent with the results. AGCRN, whose graph is learned,
reaches 5.10~mph, no better than the per-link GRU with no spatial operator (5.04~mph). Its training loss (0.225) is
clearly higher than that of the road-graph models (0.209), so the problem is underfitting rather than overfitting:
from sparse inputs and 15 days of data, the model does not find out which detectors are relevant to which links.
STID and ImputeFormer behave in the same way (training losses 0.223 and 0.226), and STAEformer in its original
configuration produced the same prediction for every link (Appendix~\ref{app:stae}). With the interpolation input,
which encodes which detectors lie near each link and in which direction, AGCRN's training loss falls to 0.198 and its
test RMSE to 4.29/4.94~mph, the level of the road-graph models. The per-link GRU, which has no spatial operator at all,
also reaches this level once it receives the interpolation input (4.29/4.95~mph), although its training loss stays
higher (0.218). The road distances contained in the interpolation are therefore sufficient, and passing messages
along the road graph adds little on top of them.

Graph neural networks that pass information along the road graph therefore cope well with sparse detectors, and the
models that fail are those that must learn spatial relations from sparse data. The road graph and the interpolation
input provide the same kind of information, namely which detectors are close to a link along the road.

\subsection{Effect of the interpolation input on each architecture}\label{sec:res-featmatch}
\begin{table}[t]
\centering
\caption{Effect of the interpolation input on every architecture (RMSE in mph on unsensored links; last column:
sensor links, next hour). ``+ interp.'': the same architecture retrained with the 12 interpolation features appended
to its per-step inputs and the interpolation baseline $b$ added to its output, so that all models in this column
receive identical inputs. $\Delta$: change due to the interpolation input. First block: the GCRN family with
different spatial operators; second block: other published architectures; last row: GCRN+ with and without
interpolation. Mean over seeds $\pm$ standard deviation where several seeds are available.}
\label{tab:featmatch}
\small
\setlength{\tabcolsep}{4pt}
\begin{adjustbox}{max width=\linewidth}
\begin{tabular}{l ccc ccc c}
\toprule
 & \multicolumn{3}{c}{Unsensored, now ($t$)} & \multicolumn{3}{c}{Unsensored, next hour ($t{+}1$)} & Sensor, $t{+}1$ \\
\cmidrule(lr){2-4}\cmidrule(lr){5-7}\cmidrule(lr){8-8}
Model & Original & + interp. & $\Delta$ & Original & + interp. & $\Delta$ & + interp. \\
\midrule
GRU (no graph) & 5.04{\scriptsize$\pm$0.01} & 4.29{\scriptsize$\pm$0.01} & -0.75 & 5.11{\scriptsize$\pm$0.01} & 4.95{\scriptsize$\pm$0.01} & -0.16 & 5.16{\scriptsize$\pm$0.02} \\
GCRN (1-hop graph) \citep{seo2018gcrn} & 4.33{\scriptsize$\pm$0.03} & 4.30{\scriptsize$\pm$0.01} & -0.03 & 4.94{\scriptsize$\pm$0.02} & 4.96{\scriptsize$\pm$0.00} & +0.02 & 5.14{\scriptsize$\pm$0.02} \\
DCRNN (kernel graph) \citep{li2018dcrnn} & 4.32{\scriptsize$\pm$0.02} & 4.30{\scriptsize$\pm$0.01} & -0.02 & 4.94{\scriptsize$\pm$0.02} & 4.92{\scriptsize$\pm$0.01} & -0.02 & 5.14{\scriptsize$\pm$0.02} \\
\midrule
AGCRN (learned graph) \citep{bai2020agcrn} & 5.10{\scriptsize$\pm$0.02} & 4.29{\scriptsize$\pm$0.01} & -0.81 & 5.13{\scriptsize$\pm$0.02} & 4.94{\scriptsize$\pm$0.02} & -0.19 & 5.16{\scriptsize$\pm$0.02} \\
IGNNK \citep{wu2021ignnk} & 4.31{\scriptsize$\pm$0.03} & 4.26{\scriptsize$\pm$0.01} & -0.06 & 4.92{\scriptsize$\pm$0.00} & 4.93{\scriptsize$\pm$0.01} & +0.00 & 5.17{\scriptsize$\pm$0.01} \\
STID \citep{shao2022stid} & 5.06{\scriptsize$\pm$0.01} & 4.31{\scriptsize$\pm$0.00} & -0.75 & 5.15{\scriptsize$\pm$0.02} & 4.97{\scriptsize$\pm$0.03} & -0.18 & 5.21{\scriptsize$\pm$0.04} \\
STAEformer \citep{liu2023staeformer} & 5.07 & 4.34 & -0.73 & 5.11 & 4.98 & -0.13 & 5.15 \\
ImputeFormer \citep{nie2024imputeformer} & 5.09{\scriptsize$\pm$0.02} & 4.31{\scriptsize$\pm$0.00} & -0.78 & 5.11{\scriptsize$\pm$0.03} & 4.96{\scriptsize$\pm$0.00} & -0.15 & 5.13{\scriptsize$\pm$0.01} \\
MoGERNN \citep{zhou2025mogernn} & 4.33{\scriptsize$\pm$0.03} & 4.26{\scriptsize$\pm$0.01} & -0.07 & 4.92{\scriptsize$\pm$0.01} & 4.92{\scriptsize$\pm$0.01} & +0.00 & 5.11{\scriptsize$\pm$0.01} \\
\midrule
GCRN+ & 4.32 & 4.27{\scriptsize$\pm$0.02} & -0.05 & 4.91 & 4.92{\scriptsize$\pm$0.00} & +0.01 & 5.13{\scriptsize$\pm$0.02} \\
\bottomrule
\end{tabular}
\end{adjustbox}
\end{table}

\begin{figure}[t]
  \centering
  \includegraphics[width=\linewidth]{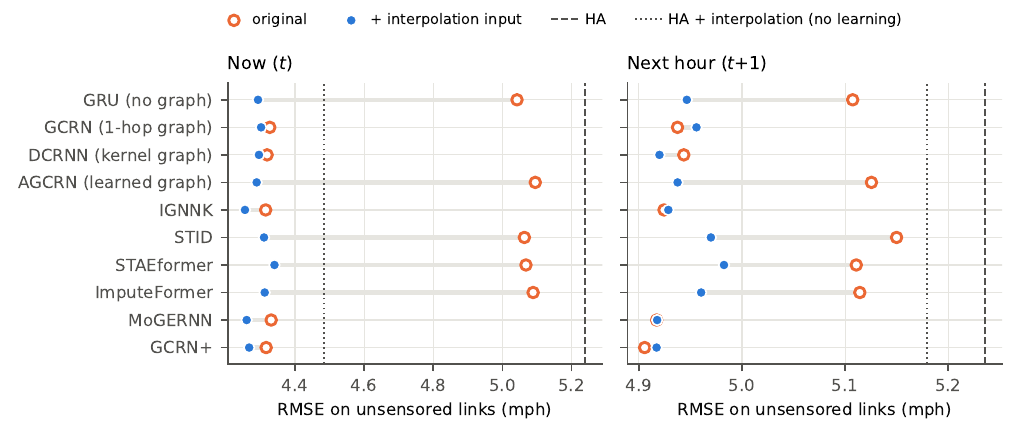}
  \caption{RMSE on unsensored links of each architecture in its original form (open circles) and with the
  interpolation input (filled circles). Vertical lines: the non-learned baselines. With the same inputs, all
  architectures are within about 0.1~mph of each other.}
  \label{fig:featmatch}
\end{figure}

Table~\ref{tab:featmatch} and Figure~\ref{fig:featmatch} show the effect of the interpolation input on every
architecture. For the per-link GRU and the four models with learned spatial structure (AGCRN, STID, STAEformer,
ImputeFormer), it lowers the current-hour RMSE by 0.73--0.81~mph (14--16\%) and the next-hour RMSE by
0.13--0.19~mph. For the models that already
use the road graph (GCRN, DCRNN, IGNNK, MoGERNN, GCRN+), the change lies between $-0.07$ and $+0.02$~mph. With the
interpolation input, the current-hour RMSE of all architectures is between 4.26 and 4.34~mph, compared with
4.31--5.10~mph without it, and the next-hour RMSE is between 4.92 and 4.98~mph. IGNNK and MoGERNN with the
interpolation input have the lowest current-hour error (4.26~mph), followed by GCRN+ (4.27~mph); GCRN+ has the lower
hourly error than IGNNK with interpolation in 42\% of the 84 test hours, and the differences are of the same size as
the variation between seeds.

When the inputs are the same, a 61k-parameter diffusion network, a Transformer with almost two million parameters and
the recurrent networks reach practically the same accuracy, and the models published in 2022--2025 do not improve on
the older graph recurrent networks. The large gaps between architectures in Table~\ref{tab:main} thus reflect the
spatial information each model receives rather than differences in temporal modelling or capacity. In practice, the
interpolation input is inexpensive (a few sparse matrix products per hour, no parameters) and makes architectures
developed for dense detector networks usable on a sparse one.

\subsection{Link-specific components}\label{sec:res-ablation}
\begin{table}[t]
\centering
\caption{Ablation of GCRN+ (RMSE in mph). All variants share the diffusion-convolutional GRU on the kernel road graph
(the backbone). ``Interpolation'' denotes the interpolation features together with the output baseline. Rows with
$\pm$: mean and standard deviation over three seeds; other rows: seed 42. Train loss as in Table~\ref{tab:why}.}
\label{tab:ablation}
\small
\begin{adjustbox}{max width=\linewidth}
\begin{tabular}{llcccc}
\toprule
Variant & Components & Unsensored $t$ & Unsensored $t{+}1$ & Sensor $t{+}1$ & Train loss \\
\midrule
Backbone (DCRNN) & kernel road graph & 4.32{\scriptsize$\pm$0.02} & 4.94{\scriptsize$\pm$0.02} & 5.10{\scriptsize$\pm$0.03} & 0.209 \\
Backbone + interp. & + interpolation input & 4.30{\scriptsize$\pm$0.01} & 4.92{\scriptsize$\pm$0.01} & 5.14{\scriptsize$\pm$0.02} & 0.206 \\
 & + interp. + link emb. & 4.25 & 4.93 & 5.12 & 0.203 \\
 & + interp. + adaptive graph & 4.30 & 4.93 & 5.10 & 0.197 \\
GCRN+ w/o interp. & + adaptive graph + link emb. & 4.32 & 4.91 & 5.09 & 0.192 \\
GCRN+ & + interp. + adaptive graph + link emb. & 4.27{\scriptsize$\pm$0.02} & 4.92{\scriptsize$\pm$0.00} & 5.13{\scriptsize$\pm$0.02} & 0.190 \\
\bottomrule
\end{tabular}
\end{adjustbox}
\end{table}

With probe labels on every link, link-specific parameters can be trained, which inductive kriging methods avoid.
Table~\ref{tab:ablation} measures their effect on top of DCRNN with the interpolation input (4.30/4.92~mph). The link
embedding lowers the current-hour RMSE to 4.25~mph, the adaptive graph leaves it unchanged, and the combination
(GCRN+) gives 4.27~mph. None of them changes the next-hour RMSE by more than about 0.01~mph. The two components lower
the training loss by 8\% (from 0.206 to 0.190) but the test error by less than 1\%. With 15 training days, the
link-specific parameters seem to fit the particular training days rather than relations that hold on new days. They
also tie the model to the links seen in training, so at this amount of data they are not worth including.

\subsection{Detector coverage}\label{sec:res-coverage}
\begin{table}[t]
\centering
\caption{Sensitivity to sensor density: RMSE (mph) on links without a sensor when only a random subset of the 643
sensor links is kept (number of sensors in parentheses). Models are retrained for each density (seed 42). Removed
sensor links join the evaluation set, so the set of evaluated links differs slightly between columns; models are
best compared within a column (best value in bold).}
\label{tab:density}
\small
\begin{adjustbox}{max width=\linewidth}
\begin{tabular}{lcccccc}
\toprule
 & \multicolumn{3}{c}{Now ($t$)} & \multicolumn{3}{c}{Next hour ($t{+}1$)} \\
Model & 100\% (643) & 50\% (322) & 25\% (161) & 100\% (643) & 50\% (322) & 25\% (161) \\
\midrule
HA & 5.24 & 5.31 & 5.33 & 5.24 & 5.30 & 5.33 \\
HA + interpolation & 4.48 & 4.71 & 4.90 & 5.18 & 5.15 & 5.28 \\
\midrule
Low-rank (PPCA) & 4.95 & 5.11 & 5.14 & 5.12 & 5.22 & 5.25 \\
\midrule
Ridge (interp.\ features) & 4.44 & 4.68 & 4.85 & 4.95 & 5.05 & 5.11 \\
GBDT (interp.\ features) & 4.36 & 4.60 & 4.80 & 4.95 & 5.08 & 5.14 \\
\midrule
GCRN (1-hop graph) & 4.33 & 4.57 & 4.78 & 4.93 & 5.03 & 5.09 \\
DCRNN (kernel graph) & 4.30 & 4.55 & 4.76 & 4.93 & \textbf{5.02} & \textbf{5.07} \\
IGNNK & 4.31 & 4.55 & 4.72 & 4.93 & 5.03 & 5.07 \\
\midrule
AGCRN (learned graph) & 5.11 & 5.16 & 5.22 & 5.15 & 5.21 & 5.26 \\
STID & 5.05 & 5.16 & 5.22 & 5.14 & 5.22 & 5.25 \\
ImputeFormer & 5.10 & 5.21 & 5.22 & 5.12 & 5.20 & 5.22 \\
\midrule
AGCRN + interp. & 4.29 & 4.53 & 4.72 & 4.96 & 5.05 & 5.09 \\
STID + interp. & 4.31 & 4.54 & 4.72 & 5.00 & 5.10 & 5.10 \\
DCRNN + interp. & 4.29 & 4.53 & 4.70 & 4.92 & 5.03 & 5.07 \\
GCRN+ & \textbf{4.25} & \textbf{4.52} & \textbf{4.69} & \textbf{4.91} & 5.03 & \textbf{5.07} \\
\bottomrule
\end{tabular}
\end{adjustbox}
\end{table}

\begin{figure}[t]
  \centering
  \includegraphics[width=\linewidth]{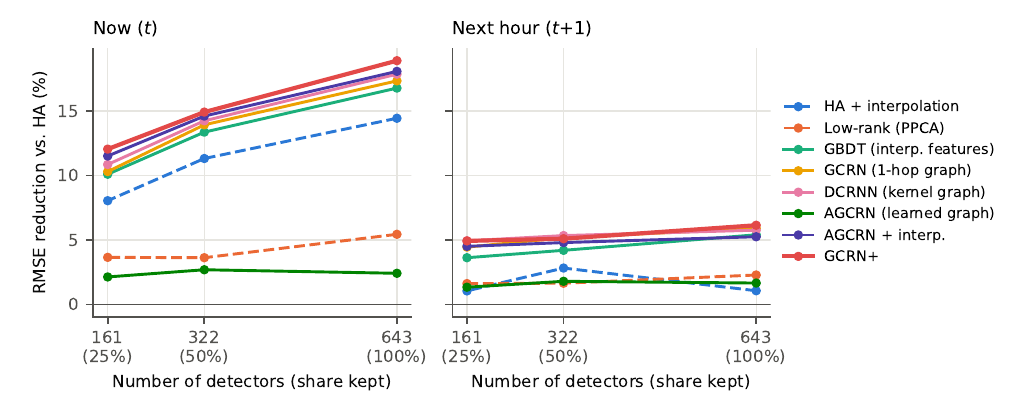}
  \caption{RMSE reduction relative to the historical average on unsensored links when the number of detectors is
  reduced (seed 42; each column is evaluated on its own set of unsensored links). Dashed: methods without a neural
  network.}
  \label{fig:density}
\end{figure}

\paragraph{Number of detectors}
Table~\ref{tab:density} and Figure~\ref{fig:density} give the results with 50\% and 25\% of the detectors. The
ranking of the families is the same at all three coverage levels. Models with road-distance information, through the
road graph or the interpolation input, stay within about 0.1~mph of each other; with 161 detectors the best are GCRN+
(4.69~mph), DCRNN + interp.\ (4.70~mph) and IGNNK, AGCRN + interp.\ and STID + interp.\ (4.72~mph), followed by DCRNN
(4.76~mph), GCRN (4.78~mph) and GBDT (4.80~mph). The models with learned spatial structure remain 2--4\% below HA at
every coverage level (AGCRN 5.11, 5.16 and 5.22~mph), and the low-rank estimator 4--6\%. For the best models, the
current-hour reduction relative to HA falls from 19\% with 643 detectors to 15\% with 322 and 12\% with 161; halving
the number of detectors thus removes about a fifth of the improvement, and keeping only a quarter removes about a
third. The interpolation input becomes slightly more useful for the road-graph models as coverage falls (for DCRNN,
0.01~mph at full coverage, 0.02~mph at 50\% and 0.06~mph at 25\%). For the next hour, the reduction relative to HA is
5--6\% at all three levels, and the interpolation baseline alone gives almost none.

\begin{table}[t]
\centering
\caption{RMSE (mph) on unsensored links by network distance to the nearest sensor (seed 42). Numbers of links
per group in the header.}
\label{tab:distance}
\small
\setlength{\tabcolsep}{3.5pt}
\begin{adjustbox}{max width=\linewidth}
\begin{tabular}{l ccccc ccccc}
\toprule
 & \multicolumn{5}{c}{Now ($t$)} & \multicolumn{5}{c}{Next hour ($t{+}1$)} \\
\cmidrule(lr){2-6}\cmidrule(lr){7-11}
Model & 0--1 mi & 1--3 mi & 3--10 mi & $>$10 mi & unreachable & 0--1 mi & 1--3 mi & 3--10 mi & $>$10 mi & unreachable \\
 & (910) & (541) & (655) & (869) & (509) & (910) & (541) & (655) & (869) & (509) \\
\midrule
HA & 5.96 & 5.61 & 5.38 & 4.28 & 4.73 & 5.98 & 5.62 & 5.38 & 4.26 & 4.69 \\
HA + interpolation & 3.27 & 5.09 & 5.38 & 4.28 & 4.73 & 5.80 & 5.59 & 5.38 & 4.26 & 4.69 \\
Covariance kriging & 5.44 & 5.50 & 5.34 & 4.20 & 4.61 & 5.86 & 5.69 & 5.46 & 4.22 & 4.69 \\
Low-rank (PPCA) & 5.32 & 5.41 & 5.23 & 4.16 & 4.65 & 5.70 & 5.55 & 5.32 & 4.18 & 4.70 \\
GBDT (interp.\ features) & 3.21 & 4.90 & 5.31 & 4.12 & 4.58 & 5.31 & 5.38 & 5.33 & 4.14 & \textbf{4.58} \\
GRU (no graph) & 5.72 & 5.46 & 5.22 & 4.10 & 4.58 & 5.81 & 5.52 & 5.29 & 4.15 & 4.60 \\
GCRN (1-hop graph) & 2.98 & 4.96 & 5.30 & 4.11 & 4.62 & 5.21 & 5.39 & 5.33 & 4.12 & 4.60 \\
DCRNN (kernel graph) & 3.06 & 4.83 & 5.24 & 4.11 & 4.60 & 5.24 & 5.35 & 5.33 & 4.14 & 4.60 \\
AGCRN (learned graph) & 5.76 & 5.51 & 5.38 & 4.11 & 4.63 & 5.82 & 5.53 & 5.40 & 4.14 & 4.65 \\
IGNNK & 2.98 & 4.85 & 5.28 & 4.12 & 4.61 & 5.21 & \textbf{5.34} & 5.33 & 4.14 & 4.64 \\
STID & 5.71 & 5.46 & 5.27 & 4.06 & \textbf{4.55} & 5.80 & 5.55 & 5.39 & 4.12 & 4.60 \\
STAEformer & 5.69 & 5.49 & 5.30 & 4.10 & 4.59 & 5.76 & 5.49 & 5.37 & 4.12 & 4.62 \\
ImputeFormer & 5.76 & 5.51 & 5.36 & 4.08 & 4.59 & 5.81 & 5.53 & 5.36 & 4.10 & 4.61 \\
MoGERNN & 3.05 & 4.85 & 5.26 & 4.10 & 4.59 & \textbf{5.21} & 5.34 & 5.29 & 4.12 & 4.59 \\
DGAE & 3.15 & 4.92 & 5.29 & 4.14 & 4.60 & 5.28 & 5.37 & 5.33 & 4.16 & 4.61 \\
MoGERNN + interp. & 2.99 & 4.84 & \textbf{5.20} & 4.07 & 4.59 & 5.23 & 5.35 & \textbf{5.25} & 4.11 & 4.59 \\
DCRNN + interp. & 2.93 & 4.85 & 5.23 & 4.12 & 4.62 & 5.22 & 5.36 & 5.29 & 4.11 & 4.62 \\
GCRN+ & \textbf{2.84} & \textbf{4.81} & 5.24 & \textbf{4.04} & 4.62 & 5.22 & 5.35 & 5.29 & \textbf{4.07} & 4.66 \\
\bottomrule
\end{tabular}
\end{adjustbox}
\end{table}

\begin{figure}[t]
  \centering
  \includegraphics[width=\linewidth]{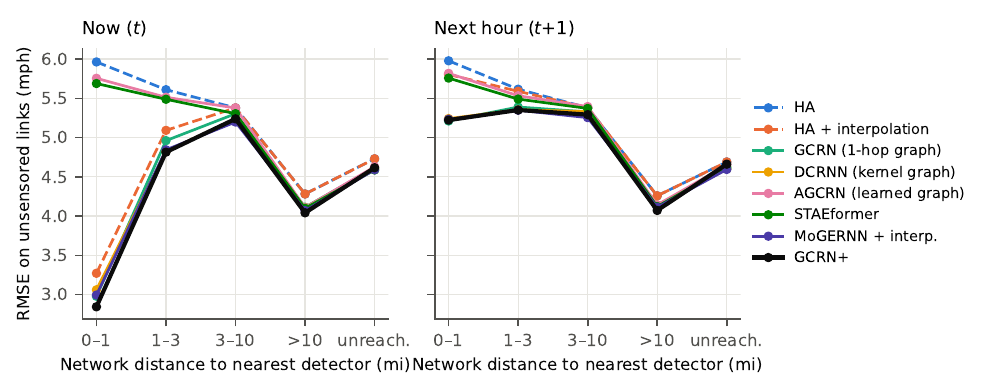}
  \caption{RMSE on unsensored links by network distance to the nearest detector (seed 42). The last group
  (unreachable) is categorical. Dashed: baselines without learning.}
  \label{fig:distance}
\end{figure}

\paragraph{Distance to the nearest detector}
Table~\ref{tab:distance} and Figure~\ref{fig:distance} break down the error at full coverage by network distance to
the nearest detector. Most of the improvement occurs close to the detectors. Within 1~mi of a detector (910 links),
the best models halve the current-hour error of HA (2.84 vs.\ 5.96~mph for GCRN+), and the interpolation baseline on
its own reaches 3.27~mph. Between 1 and 3~mi the reduction is 14\% (4.81 vs.\ 5.61~mph) and between 3 and 10~mi 3\%.
Beyond 10~mi and on links without a road-graph connection to a detector it is 2--6\%; there, the remaining
improvement comes from the regional mean anomaly, the static attributes and, for the statistical and
learned-structure models, correlations with distant detectors. For the next hour, the reduction is 13\% within 1~mi
and at most 5\% further away. Close to the detectors, where most of the information is, the models with learned
spatial structure are almost as poor as HA (AGCRN 5.76~mph within 1~mi). Overall, a detector improves estimates over
roughly 3~mi of road; beyond that, none of the models does much better than the historical profile.

\subsection{Road type, traffic volume and traffic conditions}\label{sec:res-network}
\begin{table}[t]
\centering
\caption{RMSE (mph) on unsensored links by road type and by directional AADT tercile (low $<$36k, high
$\ge$66k vehicles/day). Predictions averaged over seeds. Header: number of unsensored links and share of the
group's links that carry a sensor.}
\label{tab:network}
\small
\setlength{\tabcolsep}{3.5pt}
\begin{adjustbox}{max width=\linewidth}
\begin{tabular}{l cccccc cccccc}
\toprule
 & \multicolumn{6}{c}{Now ($t$)} & \multicolumn{6}{c}{Next hour ($t{+}1$)} \\
\cmidrule(lr){2-7}\cmidrule(lr){8-13}
 & \multicolumn{3}{c}{Road type} & \multicolumn{3}{c}{AADT} & \multicolumn{3}{c}{Road type} & \multicolumn{3}{c}{AADT} \\
\cmidrule(lr){2-4}\cmidrule(lr){5-7}\cmidrule(lr){8-10}\cmidrule(lr){11-13}
Model & Interstate & Other freeway & Tollway & Low & Medium & High & Interstate & Other freeway & Tollway & Low & Medium & High \\
 & (1721) & (1203) & (560) & (1107) & (1102) & (1111) & (1721) & (1203) & (560) & (1107) & (1102) & (1111) \\
Sensor share & 20\% & 15\% & 0\% & 11\% & 18\% & 20\% & 20\% & 15\% & 0\% & 11\% & 18\% & 20\% \\
\midrule
HA & 5.43 & 5.14 & 4.84 & 4.13 & 4.90 & 6.48 & 5.44 & 5.14 & 4.79 & 4.11 & 4.90 & 6.48 \\
HA + interpolation & 4.29 & 4.59 & 4.83 & 3.86 & 4.34 & 5.09 & 5.30 & 5.18 & 4.79 & 4.11 & 4.88 & 6.31 \\
Low-rank (PPCA) & 5.05 & 5.01 & 4.54 & 4.06 & 4.69 & 5.94 & 5.26 & 5.13 & 4.60 & 4.12 & 4.84 & 6.21 \\
GBDT (interp.\ features) & 4.18 & 4.44 & \textbf{4.49} & 3.72 & 4.16 & 4.92 & 5.04 & 4.91 & 4.57 & 3.95 & 4.64 & 5.96 \\
GCRN (1-hop graph) & 4.09 & 4.44 & 4.59 & 3.67 & 4.15 & 4.88 & 4.99 & 4.95 & 4.60 & 3.91 & 4.65 & 5.96 \\
DCRNN (kernel graph) & 4.08 & 4.42 & 4.62 & 3.65 & 4.17 & 4.87 & 5.00 & 4.94 & 4.59 & 3.90 & 4.67 & 5.97 \\
AGCRN (learned graph) & 5.21 & 5.08 & 4.57 & 3.95 & 4.77 & 6.23 & 5.24 & 5.12 & 4.56 & 3.96 & 4.80 & 6.27 \\
AGCRN + interp. & 4.01 & 4.46 & 4.56 & 3.64 & 4.18 & 4.78 & 4.96 & 4.97 & 4.58 & \textbf{3.88} & 4.67 & 5.92 \\
STAEformer & 5.20 & 5.09 & 4.58 & 3.95 & 4.77 & 6.23 & 5.26 & 5.12 & 4.61 & 3.95 & 4.82 & 6.30 \\
MoGERNN + interp. & 4.02 & \textbf{4.36} & 4.53 & 3.64 & \textbf{4.11} & 4.77 & 4.98 & \textbf{4.89} & 4.56 & 3.91 & \textbf{4.63} & 5.91 \\
GCRN+ & \textbf{3.98} & 4.44 & 4.51 & \textbf{3.63} & 4.18 & \textbf{4.72} & \textbf{4.93} & 4.95 & \textbf{4.53} & 3.89 & 4.64 & \textbf{5.88} \\
\bottomrule
\end{tabular}
\end{adjustbox}
\end{table}

Table~\ref{tab:network} breaks the errors down by road type and traffic volume. The differences between road types
mostly follow coverage. On interstates, where 20\% of the links have a detector, the best models reduce the
current-hour error of HA by 27\% (3.98 vs.\ 5.43~mph), and on other freeways (15\% coverage) by 15\%. On the tollways,
which have a single detector, the interpolation baseline does not help (4.83 vs.\ 4.84~mph) and no model gains more
than 7\% (GBDT, 4.49~mph). Here the low-rank estimator (4.54~mph) and AGCRN without interpolation (4.57~mph), which can
draw on correlations with distant detectors, do as well as the road-graph models. High-volume links have the largest
errors under HA (6.48~mph), because their congestion varies most from day to day, and also the largest reductions
(27\%, GCRN+ 4.72~mph). Low-volume links are less often instrumented (11\%) and improve by 12\%. For the next hour the
reductions are 9\% on high-volume links and on interstates and 5--6\% elsewhere.

\begin{table}[t]
\centering
\caption{RMSE (mph) on unsensored links by period (seed 42). Periods are assigned by the issue hour $t$: AM peak
6:00--9:00, PM peak 15:00--19:00. ``Congested'' = cells with a true speed below 45~mph at the target hour.}
\label{tab:period}
\small
\setlength{\tabcolsep}{4pt}
\begin{adjustbox}{max width=\linewidth}
\begin{tabular}{l cccc cccc}
\toprule
 & \multicolumn{4}{c}{Now ($t$)} & \multicolumn{4}{c}{Next hour ($t{+}1$)} \\
\cmidrule(lr){2-5}\cmidrule(lr){6-9}
Model & AM peak & PM peak & Off-peak & Congested & AM peak & PM peak & Off-peak & Congested \\
\midrule
HA & 6.11 & 6.50 & 4.59 & 12.48 & 6.13 & 6.20 & 4.70 & 12.49 \\
HA + interpolation & 5.32 & 5.56 & 3.90 & 10.49 & 6.21 & 6.30 & 4.54 & 11.86 \\
Covariance kriging & 5.95 & 6.26 & 4.38 & 11.71 & 6.20 & 6.21 & 4.65 & 12.14 \\
Low-rank (PPCA) & 5.79 & 6.15 & 4.34 & 11.86 & 5.99 & 6.07 & 4.58 & 12.20 \\
GBDT (interp.\ features) & 5.22 & 5.30 & 3.83 & \textbf{10.00} & 5.89 & 5.92 & 4.40 & \textbf{11.62} \\
GRU (no graph) & 5.88 & 6.42 & 4.37 & 12.21 & 6.01 & 6.16 & 4.55 & 12.41 \\
GCRN (1-hop graph) & 5.16 & 5.37 & 3.77 & 10.24 & 5.92 & 5.93 & 4.34 & 11.99 \\
DCRNN (kernel graph) & 5.07 & 5.34 & 3.76 & 10.33 & 6.01 & 5.89 & 4.34 & 12.05 \\
AGCRN (learned graph) & 6.12 & 6.39 & 4.42 & 12.44 & 6.13 & 6.17 & 4.56 & 12.61 \\
IGNNK & 5.19 & 5.33 & 3.73 & 10.25 & 5.93 & 5.89 & 4.35 & 12.04 \\
STID & 5.99 & 6.36 & 4.36 & 12.42 & 6.15 & 6.17 & 4.53 & 12.45 \\
STAEformer & 6.00 & 6.41 & 4.37 & 12.40 & 6.06 & 6.16 & 4.52 & 12.42 \\
ImputeFormer & 6.04 & 6.42 & 4.40 & 12.42 & 6.10 & 6.14 & 4.54 & 12.65 \\
MoGERNN & 5.08 & 5.35 & 3.76 & 10.33 & 5.83 & 5.90 & 4.35 & 11.91 \\
DGAE & 5.17 & 5.52 & 3.74 & 10.32 & 5.83 & 5.99 & 4.38 & 11.90 \\
MoGERNN + interp. & \textbf{5.03} & 5.32 & 3.73 & 10.03 & 5.84 & 5.92 & \textbf{4.33} & 11.68 \\
DCRNN + interp. & 5.12 & 5.36 & \textbf{3.71} & 10.06 & 5.89 & 5.92 & 4.33 & 11.79 \\
GCRN+ & 5.04 & \textbf{5.21} & 3.72 & 10.14 & \textbf{5.80} & \textbf{5.86} & 4.38 & 11.77 \\
\bottomrule
\end{tabular}
\end{adjustbox}
\end{table}

The errors are largest in congestion (Table~\ref{tab:period}). For cells with a true speed below 45~mph (9\% of the
unsensored current-hour cells), HA has an RMSE of 12.5~mph. The best methods reduce this to about 10.0~mph for the
current hour and to 11.6--11.8~mph for the next hour. GBDT with interpolation features is slightly better than the
others on congested cells, and the road-graph models with interpolation are best in the AM and PM peaks. Most of the
remaining error comes from non-recurrent congestion away from the detectors: a queue that forms between two
detectors several miles apart does not appear in the inputs until it reaches one of them.

\subsection{Operational products}\label{sec:app}
\begin{table}[t]
\centering
\caption{Operational products. Left: congestion detection on unsensored links (speed below 60\% of the link's
reference speed). Right: mean absolute error of the hourly network travel time index (AADT$\times$length weighted,
$\times10^{-3}$). Predictions averaged over seeds (STAEformer: one seed).}
\label{tab:app}
\small
\setlength{\tabcolsep}{4pt}
\begin{adjustbox}{max width=\linewidth}
\begin{tabular}{l ccc ccc cc}
\toprule
 & \multicolumn{3}{c}{Detection, now ($t$)} & \multicolumn{3}{c}{Detection, $t{+}1$} & \multicolumn{2}{c}{TTI MAE ($\times10^{-3}$)} \\
\cmidrule(lr){2-4}\cmidrule(lr){5-7}\cmidrule(lr){8-9}
Model & Prec. & Rec. & F1 & Prec. & Rec. & F1 & $t$ & $t{+}1$ \\
\midrule
HA & 0.778 & 0.700 & 0.737 & 0.778 & 0.699 & 0.736 & 31.6 & 42.2 \\
HA + interpolation & 0.829 & 0.762 & 0.794 & 0.767 & 0.709 & 0.737 & 23.8 & 85.1 \\
Low-rank (PPCA) & 0.790 & 0.716 & 0.751 & 0.780 & 0.705 & 0.741 & 24.6 & 36.6 \\
GBDT (interp.\ features) & 0.820 & 0.785 & 0.802 & 0.782 & 0.719 & 0.749 & 13.7 & 28.2 \\
GCRN (1-hop graph) & 0.831 & 0.786 & 0.808 & 0.792 & 0.708 & 0.748 & 17.1 & 30.6 \\
DCRNN (kernel graph) & 0.830 & 0.783 & 0.806 & 0.796 & 0.708 & 0.749 & 18.3 & 34.7 \\
DCRNN + interp. & 0.821 & 0.796 & 0.808 & 0.786 & 0.724 & 0.754 & 16.4 & 28.1 \\
AGCRN (learned graph) & 0.768 & 0.734 & 0.750 & 0.777 & 0.717 & 0.746 & 18.3 & 27.3 \\
STAEformer & 0.779 & 0.721 & 0.749 & 0.777 & 0.716 & 0.745 & 19.5 & 27.0 \\
ImputeFormer & 0.772 & 0.731 & 0.751 & 0.785 & 0.705 & 0.743 & 19.0 & 39.1 \\
MoGERNN & 0.825 & 0.784 & 0.804 & 0.790 & 0.718 & 0.752 & 17.8 & 28.8 \\
DGAE & 0.824 & 0.785 & 0.804 & 0.787 & 0.721 & 0.753 & 17.3 & 42.7 \\
MoGERNN + interp. & 0.830 & 0.789 & 0.809 & 0.789 & 0.717 & 0.752 & 15.6 & 27.8 \\
GCRN+ & 0.838 & 0.787 & 0.812 & 0.790 & 0.729 & 0.758 & 15.9 & 29.1 \\
\bottomrule
\end{tabular}
\end{adjustbox}
\end{table}

\paragraph{Congestion detection on unsensored links}
A link--hour is classified as congested when its speed is below 60\% of the link's reference speed (the 85th
percentile of its 22:00--05:00 speeds on the training days), which is the onset threshold used in the region's
congestion monitoring. The test days contain 19{,}357 congested unsensored link--hours. Table~\ref{tab:app} gives
precision, recall and F1. HA detects 70\% of the congested cells with a precision of 0.78, mostly recurrent
congestion. The interpolation baseline raises recall to 76\% and precision to 0.83. The models with road-distance
information reach a recall of 78--80\% at a precision of 0.82--0.84 (F1 0.80--0.81), which amounts to about
1{,}600--1{,}900 more congested link--hours detected over the four test days than with HA. AGCRN without
interpolation reaches an F1 of 0.75. For the next hour, all methods are only slightly better than the profile
(F1 0.74--0.76).

\begin{figure}[t]
  \centering
  \includegraphics[width=\linewidth]{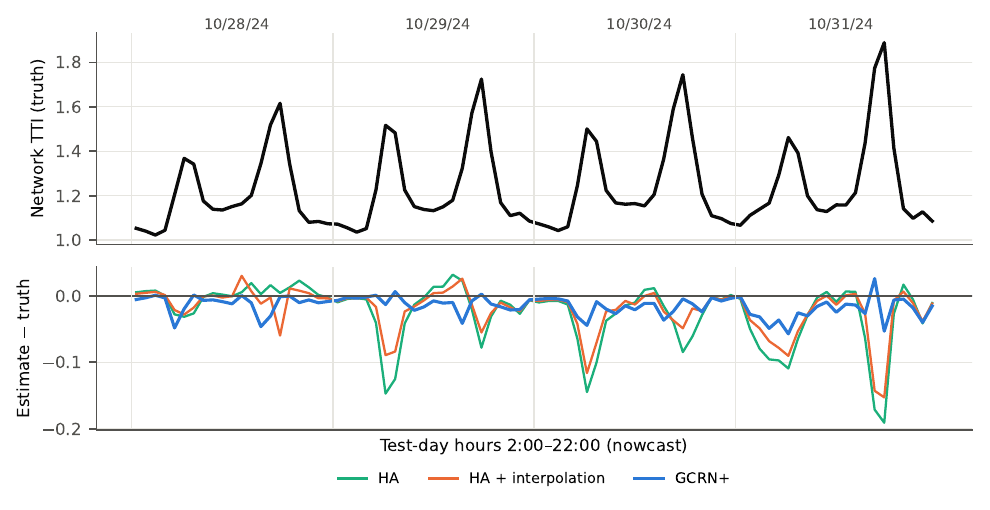}
  \caption{Network travel time index on the four test days (current hour). Top: probe ground truth. Bottom:
  estimation error when unsensored links use the historical average, HA + interpolation, or GCRN+.}
  \label{fig:tti}
\end{figure}

\paragraph{Network travel time index}
Link speeds are aggregated into an hourly network travel time index,
\begin{equation}
  \mathrm{TTI}(t)=\frac{\sum_i w_i L_i / v_{i,t}}{\sum_i w_i L_i / v^{\mathrm{ref}}_i},
\end{equation}
where $w_i$ is the directional AADT, $L_i$ the length and $v^{\mathrm{ref}}_i$ the reference speed of link $i$.
Sensor links use their readings and unsensored links the estimates. With HA on the unsensored links, the index
misses the day-to-day variation of the peaks (Figure~\ref{fig:tti}). The learned models roughly halve the mean hourly
TTI error of HA for the current hour (from $31.6\times10^{-3}$ to $14$--$20\times10^{-3}$) and reduce it by about a third
for the next hour for most models (from $42.2\times10^{-3}$ to $27$--$31\times10^{-3}$). Because link errors of opposite
sign cancel in the sum, the ranking differs from that at link level; GBDT has the lowest current-hour error
($13.7\times10^{-3}$). The interpolation baseline does well for the current hour but poorly for the next hour
($85.1\times10^{-3}$), because carrying the current anomaly of every link forward amplifies short-lived deviations
across the network. A learned model is therefore needed for next-hour products.

\begin{figure}[t]
  \centering
  \includegraphics[width=\linewidth]{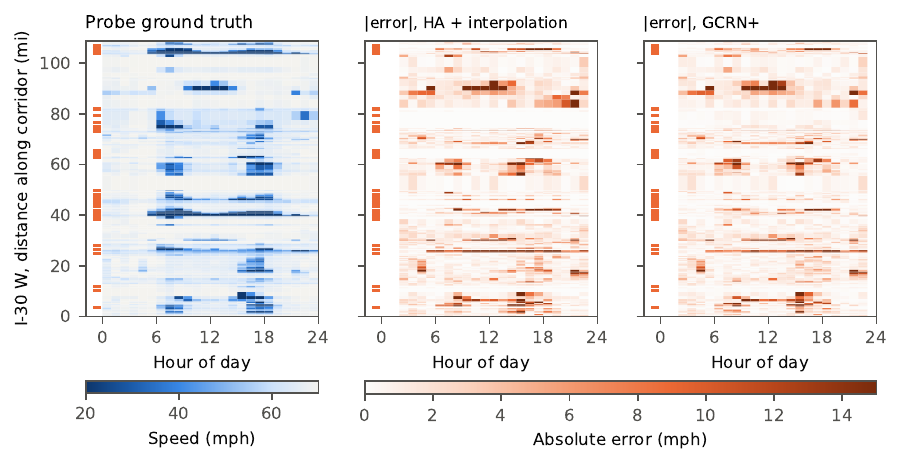}
  \caption{Time--space diagram of the corridor with the most congestion on unsensored links in the test period
  (current-hour estimates). Left: probe speeds. Middle and right: absolute error of HA + interpolation and of
  GCRN+. Orange ticks mark sensor links. Hours 0--1 (incomplete input window) and 23 (no next-hour target within the
  day) are not estimated and are left blank.}
  \label{fig:corridor}
\end{figure}

\paragraph{Corridor example}
Figure~\ref{fig:corridor} shows the corridor with the most congestion on unsensored links in the test period, I-30
westbound on 29 October 2024 (197 links over 109~mi, 38 with a detector). Interpolation reproduces congestion next to
detectors but misses most bottlenecks between distant detectors. A learned model (GCRN+ is shown) lowers the RMSE on
the corridor's unsensored links from 4.64 to 4.15~mph, mainly by estimating the extent of queues around the
detectors more accurately. Queues that do not reach a detector are missed by both.

\subsection{Recommendations}\label{sec:recommend}
Table~\ref{tab:recommend} summarises the results as guidance for agencies in a similar position. The expected
reductions refer to this network and should be taken as orders of magnitude for other networks.

\begin{table}[!htbp]
\centering
\caption{Guidance for model choice and detector placement derived from the experiments. Reduction: RMSE on
unsensored links relative to the historical average (HA), current hour / next hour.}
\label{tab:recommend}
\footnotesize
\setlength{\tabcolsep}{4pt}
\begin{tabular}{@{}>{\raggedright\arraybackslash}p{0.25\linewidth}>{\raggedright\arraybackslash}p{0.43\linewidth}
>{\raggedright\arraybackslash}p{0.13\linewidth}>{\raggedright\arraybackslash}p{0.12\linewidth}@{}}
\toprule
Situation & Recommendation & Reduction & Evidence \\
\midrule
General choice of model & Graph neural network on the road graph (e.g.\ GCRN, DCRNN, IGNNK), or any architecture
with the interpolation input & 17--19\% / 6\% & Tables~\ref{tab:main}--\ref{tab:featmatch} \\
\addlinespace[1pt]
Limited computing resources & GBDT on the interpolation features (about 0.1~mph behind the best models) & 17\% / 5\% &
Tables~\ref{tab:main}, \ref{tab:density} \\
\addlinespace[1pt]
No model training possible & HA + interpolation, for the current hour only & 14\% / 1\% & Table~\ref{tab:main} \\
\addlinespace[1pt]
Models built for dense detector networks (learned graphs, attention, identity embeddings) & Use only together with the
interpolation input & 2--4\% without, 18\% with & Tables~\ref{tab:featmatch}, \ref{tab:density} \\
\addlinespace[1pt]
About one month of probe data & Link-specific parameters are not needed; they fit the training days without improving
test accuracy & $<$1\% extra & Table~\ref{tab:ablation} \\
\addlinespace[1pt]
Half or a quarter of the current detectors & Same choice of model; expect a smaller benefit & 15\%, 12\% / 5\% &
Table~\ref{tab:density} \\
\addlinespace[1pt]
Links within 1~mi of a detector & Any of the above; interpolation alone already captures most of the gain &
52\% / 13\% & Table~\ref{tab:distance} \\
\addlinespace[1pt]
Links 1--3~mi from a detector & A learned model with road-distance information (interpolation alone: 9\%) &
14\% / 5\% & Table~\ref{tab:distance} \\
\addlinespace[1pt]
Links over 3~mi from a detector, unconnected links, uninstrumented roads (tollways) & Little is gained from any model;
GBDT or a low-rank estimator is marginally best. Priority locations for new detectors & 2--7\% / ${\le}5$\% &
Tables~\ref{tab:distance}, \ref{tab:network} \\
\addlinespace[1pt]
High-volume links & Largest errors and largest gains; good candidates for additional detectors where uncovered &
27\% / 9\% & Table~\ref{tab:network} \\
\addlinespace[1pt]
Next-hour forecasts & A learned model is needed, since carrying the current anomaly forward does not help; more gain is
likely from additional data (incidents, events, weather) than from more complex models & -- / 5--6\% &
Tables~\ref{tab:main}, \ref{tab:app} \\
\bottomrule
\end{tabular}
\end{table}

\subsection{Discussion}\label{sec:discussion}

\paragraph{Why the recent architectures do not help here}
STAEformer, ImputeFormer, STID and AGCRN were designed for networks in which every node has its own history. Their
ability to learn spatial relations from data is an advantage there, but becomes a disadvantage when 84\% of the nodes
have no current information and only 15 days of data are available. Models designed for unobserved locations (IGNNK,
MoGERNN, DGAE) aggregate over road-graph neighbours with distance weights and behave like DCRNN. With longer training
periods, the learned-structure models might discover more of the spatial relations on their own. With the amount of
data an agency typically has for a network, however, giving the model the road distances explicitly is the more
reliable option.

\paragraph{Limitations}\label{sec:limitations}
\begin{enumerate}[label=(\roman*),leftmargin=*,itemsep=2pt]
  \item One network and one month. The recommendations are based on one regional freeway network and 15 training
  days. The test set consists of four consecutive weekdays, so incidents, weather and seasonal effects are
  under-represented, and longer training periods may favour the learned-structure models and the link-specific
  components. Arterial networks, with signals and many more intersections, may behave differently. The hour-by-hour
  comparisons are descriptive; formal tests would require a day- or block-level bootstrap or a Diebold--Mariano test
  \citep{diebold1995comparing} over a longer test period.
  \item Coverage scenarios. The reduced-coverage runs remove sensor links at random and use one seed. The side-fire
  archive also contains correlated losses such as a system-wide outage (Section~\ref{sec:data}), and agencies tend to
  add or remove detectors corridor by corridor. Such losses create larger gaps and would probably reduce accuracy more
  than random removal.
  \item Temporal resolution. The models use hourly speeds. The probe data are available at 15-minute resolution, at
  which queue movements span several time steps and the recurrent part of the models may matter more.
  \item Training budget and adaptations. All models were trained with the same 40-epoch budget on a CPU, and several
  had to be adapted (Section~\ref{sec:setup}). Longer training on GPUs could change the ranking of the larger models,
  but would not by itself solve the underfitting shown in Table~\ref{tab:why}.
  \item Graph construction. Connections are inferred from endpoint proximity and turning angle rather than taken
  from a routable network with ramps. Some of the 509 links without a connection to any detector may result from
  missing interchange links.
  \item Point estimates. All models give deterministic predictions. The evidence term $c$ in
  Eq.~\eqref{eq:interp} could serve as an input to probabilistic extensions \citep{hu2023stgnp,lei2024cdste} that
  indicate how reliable each estimate is.
\end{enumerate}

\paragraph{Deployment}
A road-graph model with the interpolation input produces a network-wide estimate in about 0.15--0.2~s per hour of
data on a CPU (Table~\ref{tab:cost}). An hourly pipeline would read the last three hours of detector data, convert
the detectors' spot speeds to probe-equivalent link speeds with a site-specific relation fitted on concurrent detector
and probe data, compute the interpolation features, run the model, and publish current and next-hour speeds for every
link, from which congestion flags and the travel time index follow. Detector outages need no special treatment,
because a missing reading drops out of Eq.~\eqref{eq:interp} and the mask in the same way as in the sensor dropout used
in training. Retraining with each new batch of probe data keeps the historical profile up to date. Hours 0--1 of each
day, omitted in the experiments, can use an input window that starts on the previous day. A test on a later month with
live detector data would be the natural next step before deployment.

\section{Conclusions}\label{sec:conclusion}

Most transportation agencies have detectors on only part of their network but need speed information for all of
it. This paper compared five families of methods for estimating current and next-hour speeds from sparse fixed
detectors, with historical probe data as training labels, on the DFW freeway network (4{,}127 links, 15.6\% with a
detector), and related the results to detector coverage and road characteristics.

Graph neural networks that pass information along the road graph handled the sparse setting well. A plain GCRN on
the 1-hop adjacency was as accurate as DCRNN and as recent models designed for unobserved locations, with a
current-hour error about 17\% below the historical average on unsensored links. Neither its limited receptive field
nor the sparsity of its inputs reduced its accuracy.

Models that learn spatial relations from data (AGCRN, STAEformer, ImputeFormer, STID) stayed close to the historical
average. They did not fit the training data well either, which suggests that sparse inputs and 15 days of data are
not enough to learn which detectors matter for which links. A parameter-free kernel interpolation of the detector
readings along the road network, given as an additional input, reduced their error by 0.7--0.8~mph and brought all
ten architectures to within about 0.1~mph of each other. Link embeddings and adaptive graphs did not improve accuracy
on the test days.

Once the road distances were available to the model, accuracy depended mainly on coverage. The improvement over the
historical average was concentrated within about 3~mi of a detector, where the error was halved, and was small on
links farther away, on links with no road-graph connection to a detector and on the tollways, which have almost no
detectors. With half or a quarter of the detectors, the ranking of the methods did not change, and the
improvement of the best models over the historical average fell from 19\% to 15\% and 12\%. The models with road-distance
information detected 1{,}600--1{,}900 more congested link--hours on unsensored links than the historical average
over the four test days and halved the error of the current-hour network travel time index. Next-hour forecasts from
detector data alone improved only modestly on the historical profile.

For agencies, the results suggest that useful network-wide nowcasts can be produced from existing detectors with a
small road-graph neural network, or with gradient boosting on interpolation features, on ordinary hardware. Models
developed for dense detector networks should be given the road distances explicitly, for example through the
interpolation input. Beyond that point, additional detectors in the coverage gaps are likely to improve estimates
more than further model development (Table~\ref{tab:recommend}). Future work could use 15-minute data and longer
periods that include incidents and weather, model coverage losses by corridor and the system-wide outages seen in
the detector archive, use the same framework to rank candidate detector locations, produce probabilistic estimates from the
detector evidence term, and extend the analysis to arterial networks.

\bibliographystyle{plainnat}
\bibliography{references}

\appendix
\section{Hyperparameters}\label{app:hyper}

\begin{table}[ht]
\centering
\caption{Hyperparameters. All deep models share the training settings in the first block. ``+ interp.''\ variants
use the same settings as the original architecture.}
\label{tab:hyper}
\small
\begin{tabular}{@{}p{0.27\linewidth}p{0.69\linewidth}@{}}
\toprule
\multicolumn{2}{l}{\textit{Shared training settings}} \\
Input window / horizon & 3 hours / current + next hour \\
Optimiser & Adam, learning rate $2\times10^{-3}$ (STAEformer $10^{-3}$), halved after 2 stagnant epochs \\
Batch, epochs, early stopping & 8 windows, $\le 40$ epochs, patience 6 on validation RMSE \\
Sensor dropout & fraction $U(0, 0.4)$ of sensors hidden per sample \\
Loss & Huber ($\delta=1$) on standardised anomalies; observed current cells weight 0.1 \\
Gradient clipping & 5 \\
\midrule
\multicolumn{2}{l}{\textit{Interpolation layer}} \\
Feature kernels & $\kappa_s$ with $s\in\{1,3,8\}$~mi, downstream and upstream, truncated at $\min(3s, 15)$~mi, $\lambda=10^{-3}$ \\
Baseline & pooled kernel with $s=1$~mi, $\lambda=0.1$ (chosen on validation days) \\
\midrule
\multicolumn{2}{l}{\textit{GCRN+}} \\
Graph supports & distance kernel $\kappa_{1.5}$, truncated at 3~mi, row-normalised, downstream and upstream, $K=2$ \\
Recurrent network & 2 DCGRU layers, 32 hidden units \\
Optional components & low-rank adaptive graph, $E_1,E_2\in\mathbb R^{N\times10}$; link embedding, 8 dimensions \\
Head & MLP $(32+33)\to64\to2$ (last hidden state and last input) \\
\midrule
\multicolumn{2}{l}{\textit{Competitors}} \\
GRU / GCRN / DCRNN & 2 layers, 32 hidden, $K=2$; no graph / row-normalised 1-hop adjacency / distance-kernel supports \\
AGCRN & 2 layers, 32 hidden, embedding 10, Chebyshev order 2, low-rank adaptive graph \\
IGNNK & 3 diffusion layers, 64 hidden, $K=2$ \\
STID & embeddings 32 (series, link, time of day, attributes), 3 residual MLP layers \\
STAEformer & $d_{\text{in}}=24$, $d_{\text{tod}}=24$, $d_a=80$, attributes 16; 3+3 layers, 4 heads, FF 256, dropout 0.1, pre-LN \\
ImputeFormer & $D=64$, node embedding 80, 3 layers, 2 projection tokens, 4 heads \\
MoGERNN & 5 experts, top-3 gating, hidden sizes 32 and 16, 48 neighbours within 5~mi; 2 DCGRU layers, 32 hidden \\
Covariance kriging & empirical anomaly covariance of training days, shrinkage towards the diagonal ($\alpha=0.95$, validation) \\
Low-rank (PPCA) & rank 160, noise variance chosen on validation days \\
GBDT & histogram GBDT, 300 iterations, learning rate 0.1, 63 leaves, $4\times10^5$ training rows \\
\bottomrule
\end{tabular}
\end{table}

\begin{table}[t]
\centering
\caption{Model size and cost on a 10-core CPU. Training minutes are averaged over seeds and were measured with
four to five jobs running in parallel; they are only comparable relative to each other. For the statistical
estimators, the time covers the validation grid search and the test predictions. Inference: seconds to produce one
full test day (21 network-wide predictions).}
\label{tab:cost}
\small
\begin{adjustbox}{max width=\linewidth}
\begin{tabular}{lrrrr}
\toprule
Model & Parameters & Epochs & Train (min) & Inference (s/day) \\
\midrule
Covariance kriging & -- & -- & 17 & -- \\
Low-rank (PPCA) & -- & -- & 71 & -- \\
Ridge (interp.\ features) & -- & -- & 0 & -- \\
GBDT (interp.\ features) & -- & -- & 1 & -- \\
GRU (no graph) & 13{,}730 & 40 & 24 & 0.7 \\
GCRN (1-hop graph) & 55{,}586 & 40 & 88 & 2.0 \\
DCRNN (kernel graph) & 55{,}586 & 40 & 89 & 2.6 \\
AGCRN (learned graph) & 252{,}536 & 25 & 60 & 2.1 \\
IGNNK & 49{,}282 & 40 & 15 & 0.4 \\
STID & 233{,}026 & 37 & 19 & 0.5 \\
STAEformer & 1{,}941{,}802 & 34 & 766 & 22.1 \\
ImputeFormer & 535{,}154 & 40 & 168 & 5.6 \\
MoGERNN & 53{,}500 & 40 & 204 & 6.6 \\
DGAE & 62{,}826 & 40 & 333 & 6.1 \\
DCRNN + interp. & 62{,}114 & 33 & 85 & 3.2 \\
GCRN+ & 206{,}790 & 27 & 168 & 4.6 \\
AGCRN + interp. & 275{,}576 & 33 & 80 & 2.6 \\
IGNNK + interp. & 60{,}802 & 40 & 21 & 0.9 \\
STID + interp. & 234{,}178 & 40 & 22 & 0.7 \\
STAEformer + interp. & 1{,}942{,}090 & 30 & 692 & 23.2 \\
ImputeFormer + interp. & 535{,}922 & 27 & 206 & 5.5 \\
MoGERNN + interp. & 59{,}644 & 33 & 245 & 7.1 \\
GCRN (1-hop graph) + interp. & 62{,}114 & 35 & 79 & 4.2 \\
GRU (no graph) + interp. & 15{,}650 & 40 & 24 & 0.8 \\
\bottomrule
\end{tabular}
\end{adjustbox}
\end{table}

\FloatBarrier
\section{Collapse of STAEformer in its original configuration}\label{app:stae}

Trained with the original post-LayerNorm blocks and Xavier initialisation of the adaptive embedding (learning rate
$2\times10^{-3}$), STAEformer reached its best validation RMSE after the first epoch and stopped early after seven
epochs; its training loss stayed at 0.278 throughout. Its test RMSE on unsensored links was 5.26~mph (now) and
5.23~mph (next hour), equal to the historical average (5.24~mph) within 0.02~mph. The across-link standard deviation
of its predicted anomaly on unsensored links was $10^{-4}$, against 1.04 for the target: the model predicted
practically the same anomaly for every link. With 84\% of the tokens carrying no reading, every link attends to the
same sensor tokens, and after post-normalisation the shared attention output appears to dominate each token's own
information. The Xavier bound for a $3\times4127\times80$ embedding tensor is only $\pm0.004$, too small for the
adaptive embedding to restore link identity early in training.

With pre-LayerNorm blocks \citep{xiong2020layernorm}, $\mathcal N(0,0.1^2)$ initialisation of the adaptive
embedding and a learning rate of $10^{-3}$, the model trains normally: the across-link standard deviation of its
predicted anomaly is 0.32, and its test RMSE is 5.07/5.11~mph. This is the version reported in the main text. Its
remaining gap to the models with a distance-based spatial operator is closed by the interpolation input
(Table~\ref{tab:featmatch}).

\end{document}